\documentclass[letter,12pt]{article}

\usepackage[utf8]{inputenc}
\usepackage[T1]{fontenc}
\usepackage{amssymb}
\usepackage{amsmath}
\usepackage{stfloats}
\usepackage [autostyle, english = american]{csquotes}
\MakeOuterQuote{"}
\usepackage{subcaption}
\usepackage{textcomp}
\usepackage{float}
\usepackage{blindtext}
\usepackage{graphicx}
\usepackage{epstopdf}
\usepackage{lettrine}
\usepackage{lipsum}
\usepackage{multicol}
\usepackage{multirow}
\usepackage{textcomp}
\usepackage{layouts}
\usepackage[font=small,skip=0pt]{caption}
\usepackage{color}
\usepackage{mathtools}
\usepackage{booktabs}

\usepackage{array}
\newcolumntype{L}[1]{>{\raggedright\let\newline\\\arraybackslash\hspace{0pt}}m{#1}}
\newcolumntype{C}[1]{>{\centering\let\newline\\\arraybackslash\hspace{0pt}}m{#1}}
\newcolumntype{R}[1]{>{\raggedleft\let\newline\\\arraybackslash\hspace{0pt}}m{#1}}
\usepackage[linesnumbered]{algorithm2e}
\usepackage{bbm}
\usepackage{duckuments}
\usepackage[hidelinks]{hyperref}
\usepackage{authblk}

\usepackage{etoolbox}
\makeatletter
\def\@seccntformat#1{\@ifundefined{#1@cntformat}%
   {\csname the#1\endcsname.\hskip0.5em}    
   {\csname #1@cntformat\endcsname}
}
\appto{\appendix}{%
    \renewcommand{\appendixname}{Appendix}
    \newcommand{\section@cntformat}{\appendixname~\Alph{section}.\hskip0.5em}
    \newcommand{\subsection@cntformat}{\Alph{section}.\arabic{subsection}.\hskip0.5em}}
\makeatother

\usepackage[margin=2cm]{geometry}

\title{Sensor Fault Detection and Compensation with Performance Prescription for Robotic Manipulators}
\author[1]{S. Mohammadreza Ebrahimi}
\author[2]{Farid Norouzi}
\author[1]{Hossein Dastres}
\author[3]{Reza Faieghi\thanks{Corresponding author: reza.faieghi@torontomu.ca}}
\author[4]{Mehdi Naderi}
\author[4]{Milad Malekzadeh}
\affil[1]{Department of Electrical Engineering, Babol Noshirvani University of Technology, Babol, Iran}
\affil[2]{Department of Mechanical Engineering, Babol Noshirvani University of Technology, Babol, Iran}
\affil[3]{Department of Aerospace Engineering, Toronto Metropolitan University, Toronto, Canada}
\affil[4]{School of Production Engineering and Management, Technical University of Crete, Chania, Greece}
\date{}

\begin{document}
\maketitle

\begin{abstract}
This paper focuses on sensor fault detection and compensation for robotic manipulators.
The proposed method features a new adaptive observer and a new terminal sliding mode control law established on a second-order integral sliding surface.
The method enables sensor fault detection without the need to know the bounds on fault value and/or its derivative.
It also enables fast and fixed-time fault-tolerant control whose performance can be prescribed beforehand by defining funnel bounds on the tracking error.
The ultimate boundedness of the estimation errors for the proposed observer and the fixed-time stability of the control system are shown using Lyapunov stability analysis.
The effectiveness of the proposed method is verified using numerical simulations on two different robotic manipulators, and the results are compared with existing methods.
Our results demonstrate performance gains obtained by the proposed method compared to the existing results.
\end{abstract}

\section{Introduction}\label{se:Intro}
A fault in a robotic system can be defined as an unexpected or unplanned deviation from the normal behavior of the system, which can result in incorrect or unintended actions being taken.
One of the most common types of faults is the sensor fault which occurs when a sensor fails to accurately measure a variable or provides incorrect data.
As surveyed in \cite{amin2019review}, the common approach to detect sensor faults is to use state observers.
Examples in the context of robotic manipulators include bank of linear observers \cite{caccavale2010sensor}, $H_\infty$-based observer \cite{paviglianiti2010robust}, sliding mode observer and its higher-order variations \cite{edwards2006sensor, capisani2012manipulator}, and adaptive observer \cite{zeng2015adaptive, ma2016simultaneous}.
One notable mention is \cite{shaker2014active} in which the sensor fault detection problem is transformed into a virtual actuator fault detection framework to use the rich body of literature on actuator fault detection and compensation \cite{amin2019review}.

Once a fault is detected, fault-tolerant control (FTC) methods are required to mitigate the effects of the fault.
As pointed out in \cite{amin2019review}, one popular FTC method for nonlinear systems is sliding mode control (SMC).
Examples of using SMC in FTC include the following.
In \cite{alwi2008fault}, FTC is achieved using SMC integrated with control allocation to handle actuator faults without reconfiguring the controller.
In \cite{hamayun2011design}, an FTC scheme for over-actuated linear systems is proposed using integral SMC and control allocation techniques.
In \cite{wang2012sliding}, two SMC algorithms are presented for nonlinear systems with modeling uncertainties and actuator faults, followed by numerical verifications on a Boeing 747-100/200 model.
In \cite{shen2014integral}, two robust control methods based on integral SMC are proposed to handle actuator faults and external disturbances in spacecraft attitude dynamics.
In \cite{zeghlache2015fault}, an SMC algorithm is designed for a coaxial trirotor aircraft. This algorithm has proven to be effective in handling defects and uncertainties in the system.
In \cite{mekki2015sliding}, an SMC-based scheme for fault detection and FTC of motor systems is explored under various faults and disturbances. 
In \cite{zeghlache2018actuator}, actuator FTC is achieved for a coaxial octorotor uncrewed aerial vehicle using radial basis function networks, fuzzy logic, and SMC.
In \cite{li2022robust}, a passive SCM-based FTC is proposed for the dynamic positioning of ships in the presence of actuator faults, parameter uncertainties, and external disturbances.
In \cite{van2020self}, SMC is integrated with self-tuning proportional-integral-derivative control to achieve FTC for robotic manipulators.
Overall, SMC offers simplicity, flexibility for integration with different techniques, and robustness. 
The latter provides passive fault tolerance properties, especially for actuator faults. 
However, SMC may suffer from chattering in control input.
This has been addressed by variations of SMCs like boundary-layer SMC \cite{khalil2015nonlinear, faieghi2012novel}, fuzzy SMC \cite{faieghi2012control}, and higher-order SMC \cite{dastres2020neural}.

Most variations of SMC are designed based on asymptotic stability without a guarantee for finite convergence time.
One variation of SMC that can address this challenge is Terminal SMC (TSMC).
One of the pioneering works in TSMC is \cite{venkataraman1993control}, further expanded to multi-input and multi-output linear systems in \cite{zhihong1997terminal}, and nonlinear systems in \cite{wu1998terminal}.
A recent survey of the TSMC theory, its applications, key technical issues, and future challenges can be found in \cite{yu2020terminal}.
Recent examples of TSMC applications include the following.
In \cite{komurcugil2012adaptive}, an adaptive TSMC strategy is proposed for buck converters.
In \cite{xiong2017global}, a TSMC formulation for finite-time position and attitude tracking of a quadrotor is presented.
In \cite{boukattaya2018adaptive}, an adaptive nonsingular fast TSMC for tracking uncertain dynamical systems is developed.
In \cite{yi2019adaptive} a TSMC scheme for robotic manipulators that combines fast convergence, chattering avoidance, and adaptive estimation of uncertainties is proposed.

Although early TSMC methods guaranteed finite-time convergence, their convergence time heavily relied on the initial conditions.
The farther the initial conditions are from the desired states, the longer the convergence time is \cite{yu2020terminal, dastres2023robust, dastres2020adaptive}. 
With the emergence of performance prescription control techniques, new TSMC designs achieved fixed-time convergence irrespective of the initial conditions \cite{li2021time}.
Most fixed-time control methods suffer from singularity issues which can be partially alleviated using discontinuous control laws.
For example, in \cite{zuo2015non}, the singularity of TSMC is explained, and a new sliding surface is proposed to address it, followed by simulation-based verification on the inverted pendulum system.
Similar strategies are used in \cite{ni2016fast} for chaos suppression in power systems, in \cite{chen2018adaptive} for attitude stabilization of a spacecraft, in \cite{zhang2019fixed}, for fixed-time trajectory tracking of robotic manipulators, in \cite{chen2019adaptive}, for synchronization control of multiple robotic manipulators, and in \cite{wang2022tracking}, for constrained fractional-order nonlinear systems.
However, most existing methods rely on using the sign function in the control law which leads to chattering.
Replacing the sign function with saturation or hyperbolic tangent can resolve the chattering issue, but leads to a slower convergence rate. 
A faster chattering-free controller is developed in \cite{gao2021elm}; however, as it will be shown in our numerical simulations, this controller has a steady-state error in the presence of sensor faults.
We will build upon \cite{gao2021elm} to develop a controller with similar fast convergence characteristics but superior fault tolerance.

Overall, the objective of the present paper is to develop a fast and fixed-time sensor fault detection and compensation technique for robotic manipulators.
Inspired by \cite{shaker2014active}, we will represent sensor faults as virtual actuator faults. 
Then, we will design an adaptive observer to detect faults and estimate the actual system states.
The prominent feature of this observer is that it can guarantee the ultimate boundedness of state estimation error without the need to impose known bounds on the fault and/or its rate of change.
Furthermore, we will develop a new chattering-free fixed-time TSMC control law.
The control law lends well to the performance prescription control concept and enables imposing a boundary for the transient response of the system. 
We will integrate this control law with the aforementioned fault estimation observer and show system stability using Lyapunov stability analysis.

The main contributions of this paper include:
\begin{enumerate}
    \item Designing a new fixed time controller capable of handling sensor fault while enabling performance prescription.
    \item Developing a faster fixed-time control method inspired by \cite{gao2021elm} with a new second-order integral sliding surface. The controller offers the fast convergence property of the proposed sliding surface \cite{gao2021elm} while eliminating the steady-state error caused by perturbations, a feature that is lacking from the method in \cite{gao2021elm}.
    \item Building an adaptive observer to recover the healthy state variables from faulty measurement. The observer features a new adaptive term that makes it capable of performance guarantee without the need for the knowledge of the fault derivative.
    \item Introducing two new adaptation laws by using auxiliary variables. One allows the estimation of unknown sensor fault; while the other enables the estimation of the sensor fault derivative bound.
    \item The use of performance prescription concept to ensure errors remain in predetermined bound.
\end{enumerate}

\textbf{Notation.} Throughout the paper, vectors, and matrices, if not explicitly stated, are assumed to have appropriate dimensions.
For a given matrix $A$, $\lambda_{max}(A)$ and $\lambda_{min}(A)$ indicate the largest and smallest eigenvalues of $A$.
$O_{n \times m}$ denote $n \times m$ zero matrix and $I_n$ denotes the identity matrix.
$\| \cdot \|$ denotes the 2-norm of a vector, and $L_{\infty}$ represents the set of vectors with bounded $\infty$-norm.
\section{Problem Overview}\label{se:overview}
Let us consider the dynamics of an $n$-link robotic manipulator described as follows
\begin{equation}\label{eq:robotEq}
M\left(q\right)\ddot{q}+D\left(q,\dot{q}\right)\dot{q}+G\left(q\right)=\tau,    
\end{equation}
where $q \in \mathbb{R} ^ n$ is the joints angular position, $\tau \in \mathbb{R} ^ n$ is the vector of torques applied to each joint, $M \left( q \right) \in \mathbb{R} ^ {n \times n}$ is the inertia matrix, $D \left(q,\dot{q} \right) \in \mathbb{R} ^ {n \times n}$ is the Centrifugal and Coriolis terms matrix, and $G \left( q \right) \in \mathbb{R} ^ n$ is the gravity vector.
Equation \eqref{eq:robotEq} can be written in the following nonlinear form
\begin{equation}
    \dot{x}=Ax+H\left(x,u\right),\;
    y = Cx,
\end{equation}
where $x = (x_1, x_2)^T$ with $x_1 = q$, $x_2 = \dot q$, $u = \tau$, $y = x_1$,
\begin{equation}\label{eq:robotSS}
\begin{array}{l}
A = \left( {\begin{array}{*{20}{c}}
{{O_{n \times n}}}&{{I_n}}\\
{{O_{n \times n}}}&{{O_{n \times n}}}
\end{array}} \right),\\
H(x,u) = \left( {\begin{array}{*{20}{c}}
{{O_{n \times 1}}}\\
{{M^{ - 1}\left(x\right)}\left( {u - D(x)x_2 - G(x)} \right)}
\end{array}} \right),\; \rm{and}\\
C = \left( {{I_n},\;{O_{n \times n}}} \right).
\end{array}
\end{equation}
We assume that $H(x,u)$ is a Lipschitz function in $x$ such that
\begin{equation}\label{eq:Lipschitz}
    \left \| H(x,u)-H(\hat x,u) \right \| \le \kappa \left \| x - \hat x \right \|,
\end{equation}
where $\kappa > 0$.
This is a common assumption that has been used in observer design for robotic manipulators.
Examples include \cite{ma2016simultaneous, kang2020sensor, tran2020output}.
To verify the validity of the above inequality, one needs to calculate the Jacobian of $H\left(x,u\right)$, and $\kappa$ can be calculated as the supremum of the Jacobian over the domain of interest, as explained in \cite{yadegar2018observer}.

When a sensor fault occurs, the measurement is represented by
\begin{equation}\label{eq:y_f}
  y_f\left(t\right)=y\left(t\right)+E f\left(t\right), 
\end{equation}
where $y_f$ is the faulty measurement, and $f\in\mathbb{R}^m$ is the fault value, allowed to be time-varying. Note that we drop the time argument to simplify the mathematical notations. $E\in\mathbb{R}^{n\times m}$ is a constant matrix representing the propagation of fault in the output variable.
We assume that $f$ satisfies the following inequalities
\begin{equation}\label{eq:faultupperbounds}
  \left \| f \right \| ^ 2 \le F {\;\;\;\rm{and}\;\;\;} \left \| {\dot{f}} \right \| ^2 \le F_d,
\end{equation}
where $F$ and $F_d$ are two unknown positive constants. It is noteworthy that the above assumptions are common. 
Examples of methods that assume an upper bound on $\dot f$ include \cite{zhang2016robust} and \cite{khebbache2016adaptive}.

\begin{figure}[h]
    \centering
    \includegraphics[trim={0cm 7cm 0cm 0cm},clip,width=\linewidth]{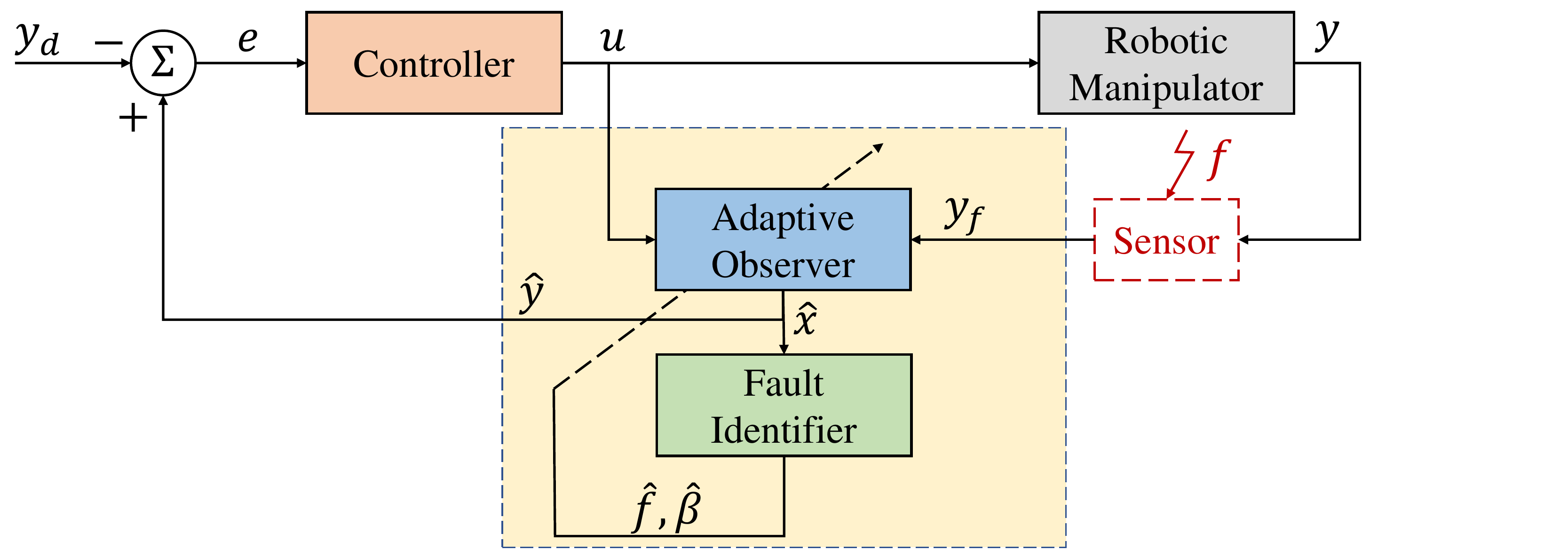}
    \caption{Block diagram of the proposed fault estimation and compensation method}
    \label{fig:blockDiagram}
\end{figure}

In this paper, we focus on both fault estimation and compensation.
For fault estimation, our objective is to design an observer that can estimate fault and also recover $y$ from faulty measurements $y_f$.
Since we assume $F$ and $F_d$ are unknown, we will use an adaptive observer to achieve the above objective.
For fault compensation, our objective is to design a fast fixed-time TSMC law that allows performance prescription on the trajectory tracking error.
This control law must use the estimated states from the observer and must guarantee the system's stability.
Figure \ref{fig:blockDiagram} illustrates an overview of the system architecture. 
In the next two sections, we will explain the details of our observer and controller design.

\section{Observer Design}\label{se:observer}
To design the observer, we first represent the sensor faults as virtual actuator faults. 
As mentioned in Section \ref{se:Intro}, this representation is inspired by \cite{shaker2014active}, and facilitates handling sensor faults. 

The sensor fault can be represented as follows,
\begin{equation}\label{eq:virtualActuator}
    {\dot{x}_v}=-A_v x_v+ A_v y_f,
\end{equation}
where $x_v \in \mathbb{R}^{n}$ is the virtual actuator state vector, and $-A_v \in \mathbb{R}^{n \times n }$ is a constant Hurwitz matrix.

Having defined sensor fault as a virtual actuator fault, an augmented system can be formed using \eqref{eq:robotSS} and \eqref{eq:virtualActuator} such that $x_a = (x, x_v)^T$ is the state vector of the augmented system with the following dynamics
\begin{equation}\label{eq:augmentedSS}
    \dot{x}_a=A_ax_a+H_a\left(x_a,u\right)+E_a f,\;
    y_a = C_a x_a = x_v,
\end{equation}
where
\begin{equation}
\begin{array}{l}
{A_a} = \left( {\begin{array}{*{20}{c}}
A&{{O_{2n \times n}}}\\
{{A_v}C}&{ - {A_v}}
\end{array}} \right),\\
{H_a (x_a,u)} = \left( {\begin{array}{*{20}{c}}
{H(x,u)}\\
{{O_{n \times 1}}}
\end{array}} \right),\\
{E_a} = \left( {\begin{array}{*{20}{c}}
{{O_{2n \times m}}}\\
{{A_v}E}
\end{array}} \right),\;{\rm{and}}\\
{C_a} = \left( {\begin{array}{*{20}{c}}
{{O_{n \times 2n}}}&{{I_n}}
\end{array}} \right).
\end{array}
\end{equation}
Note that $A_v$ must be chosen such that the pair $(A_a,C_a)$ is observable.

To design the observer, we use $\hat x_a$ to denote the estimation of $x_a$. 
Then, the estimated output will be $\hat y_a = C_a \hat x_a = \hat x_v$.
We also use the notations $\tilde x_a = x_a - \hat x_a$ and $\tilde y_a = y_a - \hat y_a$ to indicate the estimation errors.
We propose the following observer
\begin{equation}\label{eq:observer}
\dot{\hat x}_a = A_a \hat x_a + H_a \left( {\hat x_a,u} \right) + E_a \hat f\left( t \right) + L\tilde y + \Lambda \;\upsilon,
\end{equation}
where $L\in R^{3n\times n}$ and $\Lambda \in R^{3n\times n}$ are design parameters, $\hat f$ is the estimation of fault, and $\upsilon \in R^{n} $ is a term that will be introduced shortly.

Let us first explain how $\hat f$ is computed.
We represent the sensor fault as 
\begin{equation}\label{eq:fgammad}
f\left(t\right)=\gamma d\left(t\right),    
\end{equation}
where $d \in \mathbb{R}^m$, and $\gamma$ is a constant positive definite matrix in $\mathbb{R}^{m \times m}$. 
Therefore, $\hat f = \gamma \hat d$, and the problem of estimating $\hat f$ is equivalent to estimating $\hat d$. 

\textbf{Remark 1:} \textit{Note that $\gamma$ serves as a scaling factor that adds an extra degree of freedom to tune the convergence rate of the fault estimator.
While $\gamma$ can be simply set to the identity matrix in many cases, our experiences with observer design reveal that this additional degree of freedom is generally useful in system tuning.}

We use the following expressions to estimate $d$.
\begin{equation}\label{eq:dHat}
  \hat d = \hat \pi + \Gamma E^T \hat y_a,
\end{equation}
where $\hat \pi$ is an adaptive term with the following dynamics
\begin{equation}\label{eq:piAdaptRule}
    \dot{\hat \pi} = -\Gamma E^T C_a \left (A_a \hat x_a + H_a(\hat x_a,u)+E_a \hat f \right ),
\end{equation}
where $\Gamma \in \mathbb{R}^{m \times m}$ is a positive-definite design parameter. 
As it will be revealed in our Lyapunov analysis, $\hat \pi$ can be regarded as the estimate of the value represented by
\begin{equation}\label{eq:pi}
    \pi = d - \Gamma E^T y_a.
\end{equation}
The adaptation law \eqref{eq:piAdaptRule} helps the observer to deal with the unknown upper bound of $f$ i.e. $F$.

As we deal with the challenging case in which the upper bound of $\dot f$, i.e. $F_d$, is unknown too, we introduce the following expression for $\upsilon$ 
\begin{equation}\label{eq:upsilon}
\upsilon  = \Upsilon \tilde y + \hat \beta {\tilde y} {\left\| {\tilde y} \right\|}^{-2}  {{{\tanh }^2}\left( {\rho _1^{ - 1}{{\left\| {\tilde y} \right\|}^2}} \right)},
\end{equation}
where $\Upsilon \in \mathbb{R}^{n \times n}$ and is positive-definite, and $\rho_1 > 0$. 
Similar to $\hat \pi$, $\hat \beta$ is an adaptive parameter but with the following adaptation law
\begin{equation}\label{eq:betaAdaptRule}
\dot{\hat \beta}  = 2{\tanh ^2}\left( {\rho _1^{ - 1}{{\left\| {\tilde y} \right\|}^2}} \right) - \rho_2 \hat \beta,
\end{equation}
where $\rho_2 > 0$. Again, as it will be revealed in our Lyapunov analysis, $\hat \beta$ is an estimate of a parameter defined as
\begin{equation}\label{eq:beta}
    \beta  = 0.5\lambda _{\min }^{ - 1}\left( \gamma  \right){\lambda _{\max }}\left( {{{\left( {{\Gamma ^{ - 1}}} \right)}^T}{\Gamma ^{ - 1}}} \right){F_d}.
\end{equation}
Therefore, by using \eqref{eq:betaAdaptRule}, the observer can deal with unknown $F_d$.

\textbf{Remark 2:} \textit{As it will become clear shortly, the reason to use $\tanh^2\left(\cdot\right)$ is the desirable properties of this function in the vicinity of the origin. 
Our stability analysis below will lead to the evaluation of terms in the form of $1-2\tanh^2\left(\cdot\right)$ where we can apply Lemma 3 (given in Appendix A) to establish stability criteria for design parameters while ensuring the smoothness of parameters evolution with time.}

To determine the stability criteria for the design parameters, let us first write the estimation error dynamics using \eqref{eq:augmentedSS}, \eqref{eq:observer}, and $\tilde x_a = x_a - \hat x_a$.
\begin{equation}\label{eq:estimatorErrorDyn}
\begin{array}{l}
{{\dot {\tilde x}}_a} = (A_a - LC_a){\tilde x_a} + E_a\tilde f + H_a({x_a},u) - H_a({{\hat x}_a},u) - \Lambda \upsilon ,
\end{array}
\end{equation}
where $\tilde f = f - \hat f$ is the fault estimation error.
Using \eqref{eq:fgammad}, \eqref{eq:dHat}, and \eqref{eq:pi}, we have
\begin{equation}\label{eq:fTilde}
    \tilde{f} = \gamma \left(\tilde{\pi}+\Gamma E^T C_a \tilde{x_a} \right),
\end{equation}
where $\tilde \pi = \pi - \hat \pi$. Let us define the following Lyapunov function
\begin{equation}\label{eq:V1}
{V_1} = \tilde x_a^TP{\tilde x_a} + 0.5{\tilde \pi ^T}{\Gamma ^{ - 1}}\tilde \pi + 0.5\tilde{\beta}^2,
\end{equation}
where $P > 0$, and $\tilde \beta = \beta - \hat \beta$.
Taking the time derivative of $V_1$ and substituting \eqref{eq:estimatorErrorDyn} yields

\begin{equation}\label{eq:v1dot}
\begin{array}{l}
{{\dot V}_1} = \tilde{x_a}^TP{{\dot{\tilde{x}}}_a} + \dot{\tilde{x}}_a^TP{{\tilde x}_a} + {{\tilde \pi }^T}{\Gamma ^{ - 1}}\dot{\tilde{\pi}} + \tilde{\beta} \dot{\tilde{\beta}} \\
\;\;\;\;\; = \tilde x_a^T\left( {P\left( {A_a - LC_a} \right)}+{{\left( {A_a - LC_a} \right)}^T}P \right){{\tilde x}_a} + \tilde x_a^TPE_a\tilde f + {{\tilde f}^T}{E_a^T}P{{\tilde x}_a}\\
\;\;\;\;\; + \tilde x_a^TP\left( {{H_a}({x_a},u) - H({{\hat x}_a},u)} \right) + {\left( {{H_a}({x_a},u) - H_a({{\hat x}_a},u)} \right)^T}P{{\tilde x}_a}\\
\;\;\;\;\;- \tilde x_a^TP\Lambda \upsilon  - {\upsilon ^T}{\Lambda ^T}P{{\tilde x}_a}
+ {{\tilde \pi }^T}{\Gamma ^{ - 1}}\dot{\tilde{\pi}} + \tilde \beta \dot{\tilde \beta}, 
\end{array}
\end{equation}

In the following, we intend to find upper bounds on some of the terms on the right-hand side of \eqref{eq:v1dot}.
Our developments rely on the application of Young's inequality given in Appendix A.

For the second and third terms in the right-hand side of \eqref{eq:v1dot}, we apply \eqref{eq:young} to write
\begin{equation}\label{eq:2&3Term}
    \tilde x_a^TPE_a\tilde f + {\tilde f^T}{E_a^T}P{\tilde x_a}\; \le \frac{2}{3}\tilde x_a^TP{P^T}{\tilde x_a}\; + 3\;\tilde x_a^T{\varepsilon _1}\varepsilon _1^T{\tilde x_a}\; + 3\;{\tilde \pi ^T}{\varepsilon _2}\varepsilon _2^T\tilde \pi,
\end{equation}
where ${\varepsilon _1} = C_a^T{E}{\Gamma ^T}{\gamma ^T}{E_a}^T$ and ${\varepsilon _2} = \;{\gamma ^T}{E_a^T}$.
For the fourth term, we apply \eqref{eq:young} and use the Lipschitz assumption \eqref{eq:Lipschitz} to write
\begin{equation}\label{eq:4thTerm}
\begin{array}{l}
{\left( {{H_a}\left( {{x_a},u} \right) - {H_a}\left( {{{\hat x}_a},u} \right)} \right)^T}P{{\tilde x}_a}\; + \tilde x_a^TP\left( {{H_a}\left( {{x_a},u} \right) - {H_a}\left( {{{\hat x}_a},u} \right)} \right)\\
\;\;\;\;\;\;\;\;\;\;\;\;\;\;\;\;\;\;\;\;\;\;\;\; \le \frac{1}{3}\tilde x_a^TP{P^T}{{\tilde x}_a} + 3\;\kappa^2\tilde x_a^T{{\tilde x}_a}.
\end{array}
\end{equation}
For $\dot{\tilde{\pi}}$, we use \eqref{eq:augmentedSS}, \eqref{eq:piAdaptRule}, \eqref{eq:pi}, and $\tilde{\pi} = \pi - \hat{\pi}$ to write
\begin{equation}\label{eq:piErrorDyn}
\begin{array}{l}
\dot{\tilde \pi}  = \dot d - \Gamma E^T{C_a}\left( {{A_a}{{\tilde x}_a} + {E_a}\tilde f} \right) - \Gamma E^T{C_a}\left( {{H_a}\left( {{x_a},u} \right) - {H_a}\left( {{{\hat x}_a},u} \right)} \right).
\end{array}
\end{equation}
Therefore, ${{\tilde \pi }^T}{\Gamma ^{ - 1}}\dot{\tilde{\pi}}$ in \eqref{eq:v1dot} becomes
\begin{equation}\label{eq:piTildeDotInV1dot}
\begin{array}{l}
{{\tilde \pi }^T}{\Gamma ^{ - 1}}\dot{\tilde \pi} = {\tilde \pi}^T {\Gamma ^{ - 1}}\dot d - {\tilde \pi}^T E^T{C_a}{A_a}{{\tilde x}_a} - {\tilde \pi}^T E^T{C_a}{E_a}\tilde f\\
\;\;\;\;\;\;\;\;\;\;\;\;\;\;\; - {\tilde \pi}^T E^T{C_a}\left( {{H_a}\left( {{x_a},u} \right) - {H_a}\left( {{{\hat x}_a},u} \right)} \right).
\end{array}
\end{equation}
Again, by applying Young's inequality \eqref{eq:young} we can derive upper bounds for each term on the right-hand side of \eqref{eq:piTildeDotInV1dot}.
To this end, the first term leads to
\begin{equation}\label{eq:1stTermPiTildeDot}
    {\tilde \pi ^T}{\Gamma ^{ - 1}}\dot d \le \frac{1}{2}{\tilde \pi ^T}\tilde \pi  + \frac{1}{2}{\dot d^T}{\left( {{\Gamma ^{ - 1}}} \right)^T}{\Gamma ^{ - 1}}\dot d \le \frac{1}{2}{\tilde \pi ^T}\tilde \pi  + \beta,
\end{equation}
where $\beta$ was defined in \eqref{eq:beta}. Similarly, we have 
\begin{equation}\label{eq:2ndTermPiTildeDot}
    -{\tilde \pi ^T}E^T{C_a}{A_a}{\tilde x_a} \le \frac{1}{{2{c_1}}}{\tilde \pi ^T}{\varepsilon _3}\varepsilon _3^T\tilde \pi  + \frac{{{c_1}}}{2}\tilde x_a^T{\tilde x_a},
\end{equation}
where $\varepsilon_3=E^TC_aA_a$, and $c_1$ is an arbitrary positive constant.
We can also write
\begin{equation}\label{eq:3rdTermPiTildeDot}
     - {\tilde \pi ^T}E^T{C_a}{E_a}\gamma \;\Gamma {E}^T{C_a}{\tilde x_a} \le \frac{1}{{2{c_2}}}{\tilde \pi ^T}{\varepsilon _4}\varepsilon _4^T\tilde \pi  + \frac{{{c_2}}}{2}\tilde x_a^T{\tilde x_a},
\end{equation}
where $\varepsilon_4 = E^T C_a E_a \gamma \Gamma E^T C_a$ and $c_2$ is an arbitrary positive constant.
Next, we have
\begin{equation}\label{eq:4thTermPiTildeDot}
     - {\tilde \pi ^T}E^T{C_a}\left( {{H_a}\left( {{x_a},u} \right) - {H_a}\left( {{{\hat x}_a},u} \right)} \right) \le {\kappa^2}\;\tilde x_a^T{\tilde x_a} + \frac{1}{4}{\tilde \pi ^T}{\varepsilon _5}\varepsilon _5^T\tilde \pi ,
\end{equation}
where $\varepsilon_5 = E^TC_a$.
Substituting the expressions from \eqref{eq:2&3Term} to \eqref{eq:4thTermPiTildeDot} in \eqref{eq:v1dot} leads to
\begin{equation}\label{eq:v1dot2}
    {\dot V_1} \le  - \tilde x_a^T{Q_1}{\tilde x_a} - {\tilde \pi ^T}{Q_2}\tilde \pi  -  \tilde x_a^TP\Lambda \upsilon  - {\upsilon ^T}{\Lambda ^T}P{\tilde x_a} + \beta + \tilde{\beta} \dot{\tilde{\beta}},
\end{equation}
where
\begin{equation}\label{eq:Q1}
    {Q_1} =  - \left( {{{\left( {{A_a} - L{C_a}} \right)}^T}P + P\left( {{A_a} - L{C_a}} \right) + {P^2} + 4{\kappa ^2}{I_{3n}} + 3{\varepsilon _1}\varepsilon _1^T + \frac{{{c_1}}}{2} + \frac{{{c_2}}}{2}} \right),
\end{equation}
and
\begin{equation}\label{eq:Q2}
    Q_2=-\left(-E^TC_aE_a+3{\varepsilon_2\ \varepsilon}_2^T+\frac{1}{2}+\frac{1}{2c_1}{\varepsilon_3\ \varepsilon}_3^T+\frac{1}{2c_2}{\varepsilon_4\ \varepsilon}_4^T+\frac{1}{4}{\varepsilon_5\ \varepsilon}_5^T\right).
\end{equation}
Now, let us choose 
\begin{equation}\label{eq:Lambda}
    \Lambda = P^{-1}C_a^T.
\end{equation}
It follows from $\tilde{y}=C_a \tilde{x}_a$ that
\begin{equation}\label{eq:5&6Terms}
   - \tilde x_a^TP\Lambda \upsilon  - {\upsilon ^T}{\Lambda ^T}P{\tilde x_a} = -2\tilde x_a^TP\Lambda \upsilon = -2 {\tilde y}^T \upsilon.
\end{equation}
Substituting \eqref{eq:5&6Terms} in \eqref{eq:v1dot2}, adding and subtracting $2\beta \tanh^2\left(\rho_1^{-1}\left\|\tilde{y}\right\|^2\right)$, and using $\tilde {\beta} = \beta - \hat{\beta} \rightarrow \dot{\tilde{\beta}}=-\dot{\hat{\beta}}$ yields
\begin{equation}\label{eq:v1dot3}
\begin{array}{l}
{{\dot V}_1} \le  - \tilde x_a^T{Q_1}{{\tilde x}_a} - {{\tilde \pi }^T}{Q_2}\tilde \pi  - 2{{\tilde y}^T}\Upsilon \tilde y + \tilde \beta \left( 2{{{\tanh }^2}\left( {\rho _1^{ - 1}{{\left\| {\tilde y} \right\|}^2}} \right) - \dot{\hat{\beta}} } \right)\\
\;\;\;\;\;\;\;\; + \beta \left( {1 - 2{{\tanh }^2}\left( {\rho _1^{ - 1}{{\left\| {\tilde y} \right\|}^2}} \right)} \right).
\end{array}
\end{equation}
For the fourth term in the right-hand side of \eqref{eq:v1dot3}, we can use \eqref{eq:betaAdaptRule} and Young's inequality \eqref{eq:young} to write
\begin{equation}
\begin{array}{l}
    \tilde \beta \left( 2{{{\tanh }^2}\left( {\rho _1^{ - 1}{{\left\| {\tilde y} \right\|}^2}} \right) - \dot{\hat{\beta}} } \right) = \rho_2 \tilde{\beta} \hat{\beta} = \rho_2 \tilde \beta \left(\beta - \tilde{\beta} \right)\\
    \;\;\;\;\;\;\;\;\;\;\;\;\;\;\;\;\;\;\le -0.5\left(\rho_2 \tilde{\beta}^2 -\rho_2 \beta^2\right).
\end{array}
\end{equation}
Therefore,
\begin{equation}
{\dot V_1} \le  - \tilde x_a^T{Q_1}{\tilde x_a} - {\tilde \pi ^T}{Q_2}\tilde \pi  + 0.5\rho{\beta ^2} - 0.5\rho \tilde \beta ^2  + \beta \left( {1 - 2{{\tanh }^2}\left( {\rho _1^{ - 1}{{\left\| {\tilde y} \right\|}^2}} \right)} \right),
\end{equation}
which can be re-written in the form of
\begin{equation}
\dot{V}_1 \le -\alpha V_1 + \delta
\end{equation}
where $\alpha = \left\{ {\frac{{{\lambda _{\min }}\left( {{Q_1}} \right)}}{{{\lambda _{\max }}\left( P \right)}},\frac{{{2\lambda _{\min }}\left( {{Q_2}} \right)}}{{{\lambda _{\max }}\left( {{\Gamma ^{ - 1}}} \right)}},\rho } \right\}$, and
\begin{equation}\label{eq:deltaaaaaaa}
\delta  = \left\{ {\begin{array}{*{20}{c}}
{0.5{\rho _2}{\beta ^2},\;\;\;\;\;\;\;\;{\rm{if}}\left\| {\tilde y} \right\| \ge 0.8814{\rho _1},}\\
{0.5{\rho _2}{\beta ^2} + \beta ,\;\;{\rm{if}}\left\| {\tilde y} \right\| < 0.8814{\rho _1}.}
\end{array}} \right.
\end{equation}
If $Q_1 >0$, $Q_2 >0$, and $\rho_2 >0$, then from Lemma 4 (given in Appendix A) we have
\begin{equation}
    V_1(t) \le V_1(0)e^{-\alpha t} + \int_0^t e^{-\alpha(t-\tau)}\delta(\tau) d \tau.
\end{equation}
In other words, 
\begin{equation}
{V_1}\left( t \right) \le \left( {{V_1}\left( 0 \right) - \frac{\delta }{\alpha }} \right){e^{ - \alpha t}} + \frac{\delta }{\alpha }.
\end{equation}
It follows from the definition of $V_1$ \eqref{eq:V1} that 
\begin{equation}
    \left\| {{{\tilde x}_a}} \right\| \le \sqrt {\frac{{\max \left( {{V_1}\left( 0 \right),\frac{\delta }{\alpha }} \right)}}{{{\lambda _{\min }}\left( P \right)}}}.
\end{equation}
Therefore, the estimation error will remain in an adjustable neighborhood of the origin, and this guarantees that with the appropriate choice of design parameters, $\hat{x}_a$ can become close to $x_a$.


\textbf{Remark 3:} \textit{The key feature of the proposed observer \eqref{eq:observer} is the adaptation laws \eqref{eq:piAdaptRule} and \eqref{eq:betaAdaptRule} that are designed in a way that can estimate $f$ without the need to know $F$ and $F_d$. The two adaptation laws also provide a higher degree of freedom in tuning the system which can lead to improved estimation performance.} 

\section{Controller Design}\label{se:controller}
This section presents the controller design.
As mentioned in Section \ref{se:Intro}, our objective is to design a fault-tolerant controller that 
can integrate well with the fault detection observer, and establish a performance-prescribed, fast, and fixed-time convergence.

Let us first decompose $L$ and $\Lambda$ as $L=\left(L_1,L_2,L_3\right)^T$ and $\Lambda=\left(\Lambda_1,\Lambda_2,\Lambda_3\right)^T$.
Based on the observer \eqref{eq:observer}, we have
\begin{equation}\label{eq:ssForControlDesign}
    \left\{ \begin{array}{l}
{{\dot{\hat x}}_1} = {{\hat x}_2} + {L_1}\tilde y + {\Lambda _1}\;\upsilon\\
{{\dot{\hat x}}_2} = {M^{ - 1}\left(\hat x_1\right)}\left( {u - D\left( {{{\hat x}_1},{{\hat x}_2}} \right){{\hat x}_2} - G\left( {{{\hat x}_1}} \right)} \right) + {L_2}\tilde y + {\Lambda _2}\;\upsilon
\end{array} \right.
\end{equation}

Let $y_d$ be the desired trajectory, and $e = \hat x_1 - y_d$ be the tracking error. Then, the error dynamics takes the following form
\begin{equation}\label{eq:errDyn}
\left\{ {\begin{array}{*{20}{l}}
{\dot e = {{\hat x}_2} + {L_1}\tilde y + {\Lambda _1}\;v - {{\dot y}_d}}\\
{\ddot e = {M^{ - 1}\left(\hat x_1\right)}\left( {u - D\left( {{{\hat x}_1},{{\hat x}_2}} \right){{\hat x}_2} - G\left( {{{\hat x}_1}} \right)} \right) + {L_2}\tilde y}{+ {L_1}\dot{\tilde y} + {\Lambda _2}\upsilon + {\Lambda _1}\dot \upsilon - {{\ddot y}_d}}
\end{array}} \right.
\end{equation}
Using \eqref{eq:virtualActuator}, \eqref{eq:observer}, and \eqref{eq:upsilon}, we have
\begin{equation}\label{eq:ydot}
    \dot y = {\dot x_v} =  - {A_v}{x_v} + {A_v}{y_f},
\end{equation}
\begin{equation}\label{eq:yhatdot}
\dot{\hat y} = {\dot{\hat x}}_v = \left( {{A_v}C\; - {A_v}} \right){\hat{x}} + {A_v}{E}\hat f + {L_3}\tilde y + {\Lambda _3}\upsilon,
\end{equation}
and
\begin{equation}\label{eq:upsilonDot}
    \dot \upsilon = \Upsilon \dot{\tilde y} + \dot{\hat{\beta}} \tilde y{\left\| {\tilde y} \right\|^{ - 2}}{\tanh ^2}\left( {\rho _1^{ - 1}{{\left\| {\tilde y} \right\|}^2}} \right) + \hat \beta \dot{\tilde y}{\left\| {\tilde y} \right\|^{ - 2}}{\tanh ^2}\left( {\rho _1^{ - 1}{{\left\| {\tilde y} \right\|}^2}} \right) + \hat \beta \tilde y\Psi,
\end{equation}
where
\begin{equation}
    \Psi = \left\|\tilde y\right\|^{-4}\left(\Theta_1-\Theta_2\right),
\end{equation}
\begin{equation}\label{eq:theta1}
    {\Theta _1} = 2\rho _1^{ - 1}{\left\| {\tilde y} \right\|^2}\tanh \left( {\rho_1^{ - 1}{{\left\| {\tilde y} \right\|}^2}} \right)\left( {1 - {{\tanh }^2}\left( {\rho_1^{ - 1}{{\left\| {\tilde y} \right\|}^2}} \right)} \right)\left( {{\dot{\tilde y}^T}\tilde y + {{{\tilde y}}^T}{\dot{\tilde y}}} \right),
\end{equation}
and
\begin{equation}
    {\Theta _2} = {\tanh ^2}\left( {\rho_1^{ - 1}{{\left\| {\tilde y} \right\|}^2}} \right)\left( {{{\dot{\tilde y}}^T}\tilde y + {{\tilde y}^T}\dot{\tilde y}} \right).
\end{equation}

\subsection{Performance Prescription}
To enable performance prescription, let us define a prescribed performance function $\mu(t)$
as follows
\begin{equation}\label{eq:mu}
    \mu(t)=\left( \mu_0 - \mu_\infty\right)e^{-lt}+\mu_\infty,
\end{equation}
where $\mu_0 > e(0)$, and $\mu_\infty > 0$, and $l > 0$.
Note that $\mu(t)$ is a positive and descending function.
If we find a way to ensure the following criteria
\begin{equation}\label{eq:ppGoal}
    -\mu(t) \le e(t) \le \mu(t),
\end{equation}
then we can shape the tracking error trajectory using the definition of $\mu(t)$. 
As such, the inequality \eqref{eq:ppGoal} is our performance prescription control objective.

To facilitate dealing with the inequality \eqref{eq:ppGoal}, we define a new variable as follows
\begin{equation}\label{eq:zGeneral}
    z\left(t\right)=T^{-1}\left( \omega \right)
\end{equation}
where $\omega = \frac{e\left(t\right)}{\mu\left(t\right)}$, and $T\left(z\left(t\right)\right)$ is a strictly monotonic ascending function with the following properties:
\begin{equation}\label{eq:Tproperties}
    T\left(0\right)=0, \lim_{z\rightarrow+\infty}{T\left(z\right)}=+1,\lim_{z\rightarrow-\infty}{T\left(z\right)}=-1,\;{\rm{and}}\;\forall z \in L_\infty \to \left\|T(z)\right\|<1.
\end{equation}
Let $\underline z$ and $\bar z$ be the lower and upper bounds on $z$. 
Since $T$ is a strictly uniform ascending function and $\mu\left(t\right)$ is always positive, we can write
\begin{equation}    \mu\left(t\right)T\left(\underline{z}\right)\le\mu\left(t\right)T\left(z\left(t\right)\right)\le\mu\left(t\right)T\left(\overline{z}\right).
\end{equation}
According to \eqref{eq:zGeneral}, $e\left(t\right)=\mu\left(t\right)T\left(z\left(t\right)\right)$. Therefore, \eqref{eq:ppGoal} can be fulfilled by ensuring $z$ is bounded.
This implies that our performance prescription objective is to find an appropriate mapping $T$ and a control law that guarantees the boundness of $z$.

A suitable choice for the function $T$ is $\tanh \left(  \cdot  \right)$. 
Using its inverse, we have
\begin{equation}\label{eq:z}
    z = {T^{ - 1}}(\omega ) = 0.5\ln \left( {\frac{{1 + \omega }}{{1 - \omega }}} \right).
\end{equation}
As such, we will use \eqref{eq:z} in designing our TSMC control law.

\subsection{Terminal Sliding Mode Dynamics}\label{se:tsmdsubsection}
Our TSMC developments here are inspired by \cite{gao2021elm}.
The TSMC approach presented in \cite{gao2021elm} has the advantage of fast fixed-time convergence; however, as it will be shown in our numerical simulations, it suffers from steady-state error in the presence of sensor fault. 
To address this, we propose a new integral second-order sliding surface, as opposed to the first-order sliding surface developed in \cite{gao2021elm}.

\textbf{Remark 4:} \textit{As it will be shown in our theoretical developments and numerical simulations, using the new integral second-order sliding surface reduces the steady-state error while maintaining similar fast convergence characteristics of its first-order counterpart in \cite{gao2021elm}.}

\textbf{Remark 5:} \textit{Note that we will base our developments on $z$ instead of $e$ to enable performance prescription.
This is another key difference of our work compared to \cite{gao2021elm}.}

Let us define the following sliding surface
\begin{equation}\label{eq:sigma}
\sigma  = \dot \eta  + \bar g\left( \eta  \right),
\end{equation}
where
\begin{equation}\label{eq:eta}
    \eta=z+\int_{0}^{t}\bar{g}\left(z\left(\tau\right)\right)d\tau.
\end{equation}
In this definition, $g\left(\cdot\right)$ is a scalar function defined as follows
\begin{equation}\label{eq:g}
    g\left(\chi\right)=\frac{1}{\varphi\left(\chi\right)}\left(\underline{\lambda}\ {\mathop{\rm sgn}}\left(\chi\right)\left|\chi\right|^{p^\ast}+\bar{\lambda}\ {\mathop{\rm sgn}}\left(\chi\right)\left|\chi\right|^\frac{\bar{p}}{\bar{q}}\right),
\end{equation}
where $\chi$ is an arbitrary scalar serving as the argument for $g$, and 
\begin{equation}
    p^\ast=\frac{1}{2}+\frac{\underline{p}}{2\underline{q}}+\left(\frac{\underline{p}}{2\underline{q}}-\frac{1}{2}\right){\mathop{\rm sgn}} \left(\left|\chi\right|-1\right),
\end{equation}
\begin{equation}
    \varphi\left(\chi\right)=a+\left(1-a\right)e^{-b\left|\chi\right|^c},
\end{equation}
$\underline \lambda$, $\bar \lambda$, $a < 1$, and $b$ are positive constants, $c$ is an even positive integer, $\underline p$, $\underline q$, $\bar p$, and $\bar q$ are positive odd integers such tat $\underline p > \underline q$ and $\bar p < \bar q$.
To construct the sliding surface, we define two vectors as $\bar g\left( z \right) = {\left( {g\left( {{z_1}} \right),g\left( {{z_2}} \right), \cdots ,g\left( {{z_n}} \right)} \right)^T}$ and $\bar g\left( \eta \right) = {\left( {g\left( {{\eta_1}} \right),g\left( {{\eta_2}} \right), \cdots ,g\left( {{\eta_n}} \right)} \right)^T}$.

As it will be revealed in our Lyapunov analysis shortly, the use of \eqref{eq:sigma} will lead to the following terminal sliding dynamics
\begin{equation}\label{eq:TSMDyn}
\left\{ \begin{array}{l}
{\chi _2} = {\chi _1} + \int_0^t {\frac{1}{{\varphi \left( {{\chi _1}\left( \tau  \right)} \right)}}(\;{\underline \lambda \mathop{\rm sgn}} \left( {{\chi _1}\left( \tau  \right)} \right){{\left| {{\chi _1}\left( \tau  \right)} \right|}^{{p^ * }}} + \bar \lambda \;{\mathop{\rm sgn}} ({\chi _1}(\tau )){{\left| {{\chi _1}(\tau )} \right|}^{\frac{{\bar p}}{{\bar q}}}})d\tau } ,\\
{{\dot \chi }_2} =  - \frac{1}{{\varphi \left( {{\chi _2}} \right)}}(\;{\underline \lambda \mathop{\rm sgn}} \left( {{\chi _2}} \right){\left| {{\chi _2}} \right|^{{p^ * }}} + \bar \lambda \;{\mathop{\rm sgn}} ({\chi _2}){\left| {{\chi _2}} \right|^{\frac{{\bar p}}{{\bar q}}}}),
\end{array} \right.
\end{equation}
where $\chi_1 = z_i$ and $\chi_2 = \eta_i$ for $i = 1, \cdots , n$.
We will establish our convergence results based on findings in \cite{gao2021elm} which is summarized as Lemma 6 in Appendix A of this paper.

First, let us take the derivative of the first equation in \eqref{eq:TSMDyn}.
We have
\begin{equation}\label{eq:TSMDynDerivative}
    \dot{\chi}_2=\dot{\chi}_1+\frac{1}{\varphi\left(\chi_1\right)}\left(\underline{\lambda}\ {\rm{sgn}}\left(\chi_1\right)\left|\chi_1\right|^{p^\ast}+\bar{\lambda}{\rm{sgn}}\left(\chi_1\right)\left|\chi_1\right|^\frac{\bar{p}}{\bar{q}}\right).
\end{equation}

According to Lemma 6 and the second equation of \eqref{eq:TSMDyn}, $\chi_2$ will converge to the origin in fixed time.
When $\chi_2 = 0$ and $\dot \chi_2 = 0$, \eqref{eq:TSMDynDerivative} becomes
\begin{equation}
    \dot{\chi_1}=-\frac{1}{\varphi\left(\chi_1\right)}\left(\underline{\lambda}\ {\rm{sgn}}\left(\chi_1\right)\left|\chi_1\right|^{p^\ast}+\bar{\lambda}\ {\rm{sgn}}\left(\chi_1\right)\left|\chi_1\right|^\frac{\bar{p}}{\bar{q}}\right).
\end{equation}
Again, based on Lemma 6, we infer that $\chi_1$ will converge to the origin in fixed time.

Note that in the definition of the function $g$ and its design parameters, we follow the criteria given by Lemma 6.
Since $\dot \chi_1$ and $\dot \chi_2$ have the same dynamics as in \eqref{eq:lemma6-1}, the convergence time for them will be in the form of \eqref{eq:convergenceTime}, which is identical to \cite{gao2021elm}.

\subsection{Control Law and Stability Analysis}
This section presents our fault-tolerant control law based on the sliding surface \eqref{eq:sigma}.

Similar to most sliding mode control designs, let us consider the Lyapunov function
\begin{equation}\label{eq:v2}
    V_2=\frac{1}{2}\sigma^T\sigma,
\end{equation}
Taking the time-derivative of \eqref{eq:v2} yields $\dot V_2 = \sigma ^T {\dot \sigma}$. In the following, we calculate $\dot \sigma$ and substitute in $\dot V_2$.

Evaluating \eqref{eq:sigma} leads to the following expression
\begin{equation}\label{eq:sigmaDot1}
\dot{\sigma}=\ddot{z}+\dot{\bar g}\left(z\right)+\dot{\bar g}\left(\eta\right).
\end{equation}

For the first term in the right-hand side of \eqref{eq:sigmaDot1}, we take the derivative of \eqref{eq:z}.
It follows that 
\begin{equation}\label{eq:zdot}
\dot z = \frac{{\partial {T^{ - 1}}}}{{\partial \omega }}\dot \omega  = r\left( {\dot e - se} \right),
\end{equation}
where 
\begin{equation}
r = {\rm{diag}}\left( {{{\left( {\mu \;\left( {1 - {\omega _1}^2} \right)} \right)}^{ - 1}}, \ldots ,{{\left( {\mu \;\left( {1 - {\omega _n}^2} \right)} \right)}^{ - 1}}} \right),
\end{equation}
and $s = \dot \mu \mu^{-1} I_n$.
Next,
\begin{equation}
    \ddot{z}=\dot{r}\left(\dot{e}-se\right)+r\left(\ddot{e}-\dot{s}e-s\dot{e}\right),
\end{equation}
where
\begin{equation}
    \dot s = {\rm{diag}}\left( {\frac{{\ddot \mu \;\mu  - {{\dot \mu }^2}}}{{{\mu ^2}}}, \ldots ,\frac{{\ddot \mu \;\mu  - {{\dot \mu }^2}}}{{{\mu ^2}}}} \right),
\end{equation}
and
\begin{equation}
    \dot{r}={\rm{diag}}\left(\frac{-\dot{\mu}\left(1-{\omega_1}^2\right)+2\ \mu\omega_1{\dot{\omega}}_1}{\left(\mu\ \left(1-{\omega_1}^2\right)\right)^2},\ldots,\frac{-\dot{\mu}\ \left(1-{\omega_n}^2\right)+2\mu\ \omega_n{\dot{\omega}}_n}{\left(\mu\ \left(1-{\omega_n}^2\right)\right)^2}\right).
\end{equation}
By defining $R = \dot r\left( {\dot e - se} \right) - r\left( {\dot se + s\dot e} \right)$, $\ddot{z}$ can be written as
\begin{equation}\label{eq:z2dot}
    \ddot{z}=R+r\ddot{e}.
\end{equation}

For the second and third terms in the right-hand side of \eqref{eq:sigmaDot1}, we take the derivative of \eqref{eq:g}.
This leads to the following expression
\begin{equation}\label{eq:gdot}
    \begin{array}{l}
\dot g\left( \chi  \right) =  - \frac{{\dot \varphi \left( \chi  \right)}}{{{\varphi ^2}\left( \chi  \right)}}\left( {\;{\underline \lambda \mathop{\rm sgn}} \left( \chi  \right){{\left| \chi  \right|}^{{p^ * }}} + \bar \lambda \;{\mathop{\rm sgn}} \left( \chi  \right){{\left| \chi  \right|}^{\frac{{\bar p}}{{\bar q}}}}} \right)\\
\;\;\;\;\;\;\;\;\;\;\; + \frac{1}{{\varphi \left( \chi  \right)}}\left(\underline \lambda {{p^ * }{{\left| \chi  \right|}^{{p^ * } - 1}} + \bar \lambda \frac{{\bar p}}{{\bar q}}{{\left| \chi  \right|}^{\frac{{\bar p}}{{\bar q}} - 1}}} \right)\dot \chi,
\end{array}
\end{equation}
where
\begin{equation}
    \dot \varphi \left( \chi  \right) =  - \left( {1 - a} \right)bc{\mathop{\rm sgn}} \left( \chi  \right){\left| \chi  \right|^{c - 1}}\dot \chi {e^{ - b{{\left| \chi  \right|}^c}}}.
\end{equation}
Substituting \eqref{eq:z2dot} and \eqref{eq:gdot} into \eqref{eq:sigmaDot1} yields
\begin{equation}\label{eq:sigmaDot2}
\dot{\sigma}=\ddot{z}+\dot{g}\left(z\right)+\dot{g}\left(\eta\right)=R+r\ddot{e}+\dot{g}\left(z\right)+\dot{g}\left(\eta\right).
\end{equation}
Then, from \eqref{eq:errDyn} it follows that
\begin{equation}\label{eq:sigmdaDot3}
\begin{array}{l}
\dot \sigma  = r{M^{ - 1}\left(\hat x_1\right)}\left( {u - D\left( {{{\hat x}_1},{{\hat x}_2}} \right){{\hat x}_2} - G\left( {{{\hat x}_1}} \right)} \right)\;\\
\;\;\;\;\;\; + M\left( {{{\hat x}_1}} \right)\left( {{L_2}\tilde y + {L_1}\dot{\tilde y} + {\Lambda _2}\;v + {\Lambda _1}\;\dot v - {{\ddot y}_d}} \right)\\
\;\;\;\;\;\; + R + \dot g\left( z \right) + \dot g\left( \eta  \right).
\end{array}
\end{equation}
Let us define the control law as follows
\begin{equation}\label{eq:u}
\begin{array}{l}
u = D\left( {{{\hat x}_1},{{\hat x}_2}} \right)\;{{\hat x}_2} + G\left( {{{\hat x}_1}} \right) - M\left(\hat x_1\right)\;\left( {{L_2}\tilde y + {L_1}\dot {\tilde y} + {\Lambda _2}\;v + {\Lambda _1}\;\dot v - {{\ddot y}_d}} \right)\\
\;\;\; - M\left(\hat x_1\right)r^{-1}\left( R + \dot g\left( z \right) + \dot g\left( \eta  \right) + {k_{1\sigma }}\sigma  + c_{1\sigma}{\sigma^{{p_\sigma} \mathord{\left/
 {\vphantom {{{p_\sigma }} {{q_\sigma }}}} \right.
 \kern-\nulldelimiterspace} {{q_\sigma }}}} + {{\bar c}_{1\sigma }}{\sigma ^{{{{{\bar p}_\sigma }} \mathord{\left/
 {\vphantom {{{{\bar p}_\sigma }} {{{\bar q}_\sigma }}}} \right.
 \kern-\nulldelimiterspace} {{{\bar q}_\sigma }}}}}\right),
\end{array}
\end{equation}
where $k_{1\sigma}$ , $c_{1\sigma}$ and ${\bar{c}}_{1\sigma}$ are positive constants, and $p_\sigma$, $q_\sigma$, ${\bar{p}}_\sigma$ and ${\bar{q}}_\sigma$ are positive odd numbers such that $p_\sigma > q_\sigma$ and ${\bar{p}}_\sigma<{\bar{q}}_\sigma$. Furthermore, $\sigma^{p_\sigma/q_\sigma}=\left({\sigma_1}^{p_\sigma/q_\sigma},\ldots,{\sigma_n}^{p_\sigma/q_\sigma}\right)^T$ and $\sigma^{{\bar{p}}_\sigma/{\bar{q}}_\sigma}=\left({\sigma_1}^{{\bar{p}}_\sigma/{\bar{q}}_\sigma},\ldots,{\sigma_n}^{{\bar{p}}_\sigma/{\bar{q}}_\sigma}\right)^T$. 

Substituting \eqref{eq:u} in \eqref{eq:sigmdaDot3} and using $\dot V_2 = \sigma ^T {\dot{\sigma}}$ yields
\begin{equation}
{\dot V_2} =  - {k_{1\sigma }}{\sigma ^T}\sigma  - {c_{1\sigma }}{\sigma ^T}{\sigma ^{{{{p_\sigma }} \mathord{\left/
 {\vphantom {{{p_\sigma }} {{q_\sigma }}}} \right.
 \kern-\nulldelimiterspace} {{q_\sigma }}}}} - {\bar c_{1\sigma }}{\sigma ^T}{\sigma ^{{{{{\bar p}_\sigma }} \mathord{\left/
 {\vphantom {{{{\bar p}_\sigma }} {{{\bar q}_\sigma }}}} \right.
 \kern-\nulldelimiterspace} {{{\bar q}_\sigma }}}}},
 \end{equation}
where
\begin{equation}
{\sigma ^T}{\sigma ^{{{{p_\sigma }} \mathord{\left/
 {\vphantom {{{p_\sigma }} {{q_\sigma }}}} \right.
 \kern-\nulldelimiterspace} {{q_\sigma }}}}} = {\sigma _1}^{1 + {p_\sigma }/{q_\sigma }} +  \ldots  + {\sigma _n}^{1 + {p_\sigma }/{q_\sigma }},
 \end{equation}
and
\begin{equation}
{\sigma ^T}{\sigma ^{{{{{\bar p}_\sigma }} \mathord{\left/
 {\vphantom {{{{\bar p}_\sigma }} {{{\bar q}_\sigma }}}} \right.
 \kern-\nulldelimiterspace} {{{\bar q}_\sigma }}}}} = {\sigma _1}^{1 + {{{{\bar p}_\sigma }} \mathord{\left/
 {\vphantom {{{{\bar p}_\sigma }} {{{\bar q}_\sigma }}}} \right.
 \kern-\nulldelimiterspace} {{{\bar q}_\sigma }}}} +  \ldots  + {\sigma _n}^{1 + {{{{\bar p}_\sigma }} \mathord{\left/
 {\vphantom {{{{\bar p}_\sigma }} {{{\bar q}_\sigma }}}} \right.
 \kern-\nulldelimiterspace} {{{\bar q}_\sigma }}}}.
 \end{equation}
Based on the definition of $V_2$ in \eqref{eq:v2}, \eqref{eq:lemma2-1} and \eqref{eq:lemma2-2}, we have
\begin{equation}
{\alpha _\sigma }V_2^{{\beta _\upsilon }} \le {\sigma ^T}{\sigma ^{{{{p_\sigma }} \mathord{\left/
 {\vphantom {{{p_\sigma }} {{q_\sigma }}}} \right.
 \kern-\nulldelimiterspace} {{q_\sigma }}}}}
 \end{equation}
where 
${\beta _\upsilon } = {{\left( {\;{p_\sigma } + {q_\sigma }} \right)} \mathord{\left/
 {\vphantom {{\left( {\;{p_\sigma } + {q_\sigma }} \right)} {2{q_\sigma }}}} \right.
 \kern-\nulldelimiterspace} {2{q_\sigma }}}$,  ${\alpha _\sigma } = {2^{{{\left( {\;{p_\sigma } + {q_\sigma }} \right)} \mathord{\left/
 {\vphantom {{\left( {\;{p_\sigma } + {q_\sigma }} \right)} {2{q_\sigma }}}} \right.
 \kern-\nulldelimiterspace} {2{q_\sigma }}}}}\;{n^{{{\left( {\;{p_\sigma } - {q_\sigma }} \right)} \mathord{\left/
 {\vphantom {{\left( {\;{p_\sigma } - {q_\sigma }} \right)} {2{q_\sigma }}}} \right.
 \kern-\nulldelimiterspace} {2{q_\sigma }}}}}$, and
\begin{equation}
{{\bar \alpha }_\sigma }V_2^{{{\bar \beta }_\upsilon }} \le {\sigma ^T}{\sigma ^{{{{{\bar p}_\sigma }} \mathord{\left/
 {\vphantom {{{{\bar p}_\sigma }} {{{\bar q}_\sigma }}}} \right.
 \kern-\nulldelimiterspace} {{{\bar q}_\sigma }}}}},
 \end{equation}
with ${\bar{\beta}}_\upsilon=\frac{\left(\ {\bar{p}}_\sigma+{\bar{q}}_\sigma\right)}{2{\bar{q}}_\sigma}$ and ${\bar{\alpha}}_\sigma=2^\frac{\left(\ {\bar{p}}_\sigma+{\bar{q}}_\sigma\right)}{2{\bar{q}}_\sigma}$.
Thus, $\dot V_2$ satisfies
\begin{equation}
    \dot{V}_2\le-2k_{1\sigma}\ V_2-c_{1\sigma}\alpha_\sigma\ V_2^{\beta_\upsilon}-{\bar{c}}_{1\sigma}{\bar{\alpha}}_\sigma\ V_2^{{\bar{\beta}}_\upsilon}.
\end{equation}
Let $\alpha_\nu=c_{1\sigma}\alpha_\sigma$ and ${\bar{\alpha}}_\nu={\bar{c}}_{1\sigma}{\bar{\alpha}}_\sigma$. Then, it follows that
\begin{equation}
    \dot{V}_2\le-\alpha_\nu V_2^{\beta_\upsilon}-{\bar{\alpha}}_\nu V_2^{{\bar{\beta}}_\upsilon},
\end{equation}
which guarantees fixed-time stability of $\sigma$ according to Lemma 5 (given in Appendix A).
This implies that $\eta$ and $z$ are fixed-time stable.

Furthermore, based on \eqref{eq:z}, we infer that $e \rightarrow 0$  when $z \rightarrow 0$.
Since $\tilde x_1 = x_1 - \hat x_1$, we can write $x_1 = \tilde x_1 + \hat x_1 = \tilde x_1 + e + y_d$. 
Since $e \rightarrow 0$ and $\tilde x_1$ is ultimately bounded, it can be inferred that $x_1$ converges to a neighborhood of $y_d$.

We can also show that $x_2$ converges to a neighborhood of $\dot y_d$. 
To this end, from \eqref{eq:sigma} and \eqref{eq:eta}, we note that $\sigma = \dot z + \bar g\left(z\right) + \bar g\left(\eta\right)$. 
Since $\bar g(0) = 0$, at the time that $\sigma \rightarrow 0$, we have ${\dot z} \rightarrow 0$.
It follows from \eqref{eq:zdot} that $\dot e \rightarrow e$.
When $\sigma \rightarrow 0$, we have $e \rightarrow 0$.
Therefore, $\dot e \rightarrow 0$.

With that in mind, we can now examine the first equation in \eqref{eq:errDyn} as follows.
Since the observer guarantees the ultimate boundedness of estimation error, $\tilde y$ is arbitrarily small. 
This implies that $\upsilon$ is arbitrarily small as it is given by \eqref{eq:upsilon}.
We also established that $\dot e \rightarrow 0$ in the above paragraph.
Putting all these together and evaluating the first equation in \eqref{eq:errDyn} implies that ${\hat x}_2$ will converge to a neighborhood of $\dot{y}_d$.
Finally, we note that $x_2 = {\tilde x}_2$ + ${\hat x}_2$, and since ${\tilde x}_2$ is arbitrarily small, we can infer that $x_2$ will converge to a neighborhood of ${\dot y}_d$.

\section{Numerical Simulations}\label{se:Simulations}
This section presents numerical simulations to evaluate the effectiveness of the proposed method.

\subsection{Example 1: Two-Link Rigid Robotic Manipulator}
\begin{figure}[h]
    \centering
    \includegraphics[trim={9cm 0cm 4.5cm 1cm},clip,scale = 0.3]{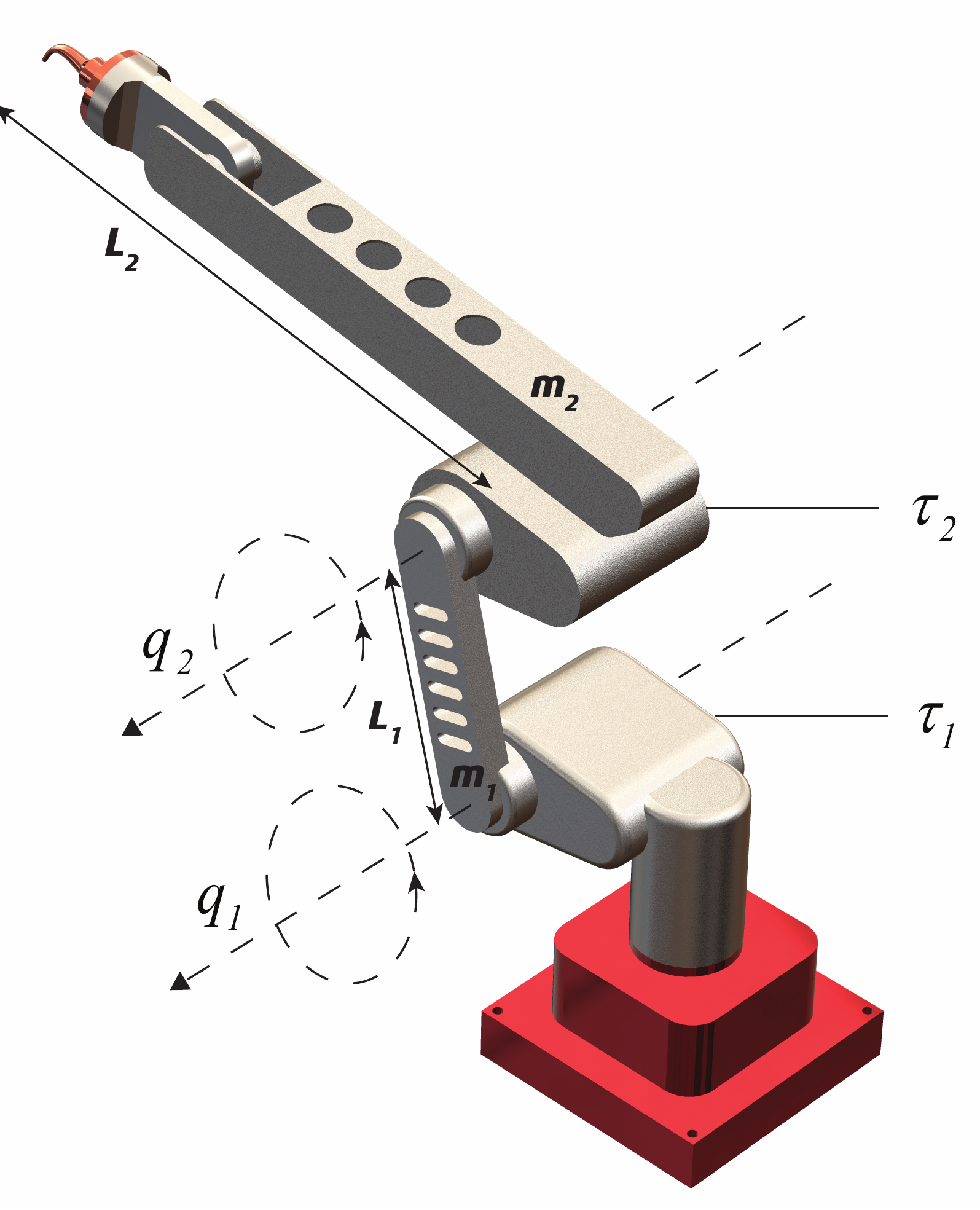}
    \caption{Two-link rigid robotics manipulator system}
    \label{fig:2linkrobot}
\end{figure}
In this example, we consider a two-link rigid robotic manipulator (Fig. \ref{fig:2linkrobot}), and compare our method with \cite{gao2021elm} and \cite{ma2016simultaneous}. 
The parameters of the robot model are selected according to \cite{gao2021elm} to minimize confounding errors in comparisons.

\begin{figure}[t]
    \centering
    \includegraphics[width = \linewidth]{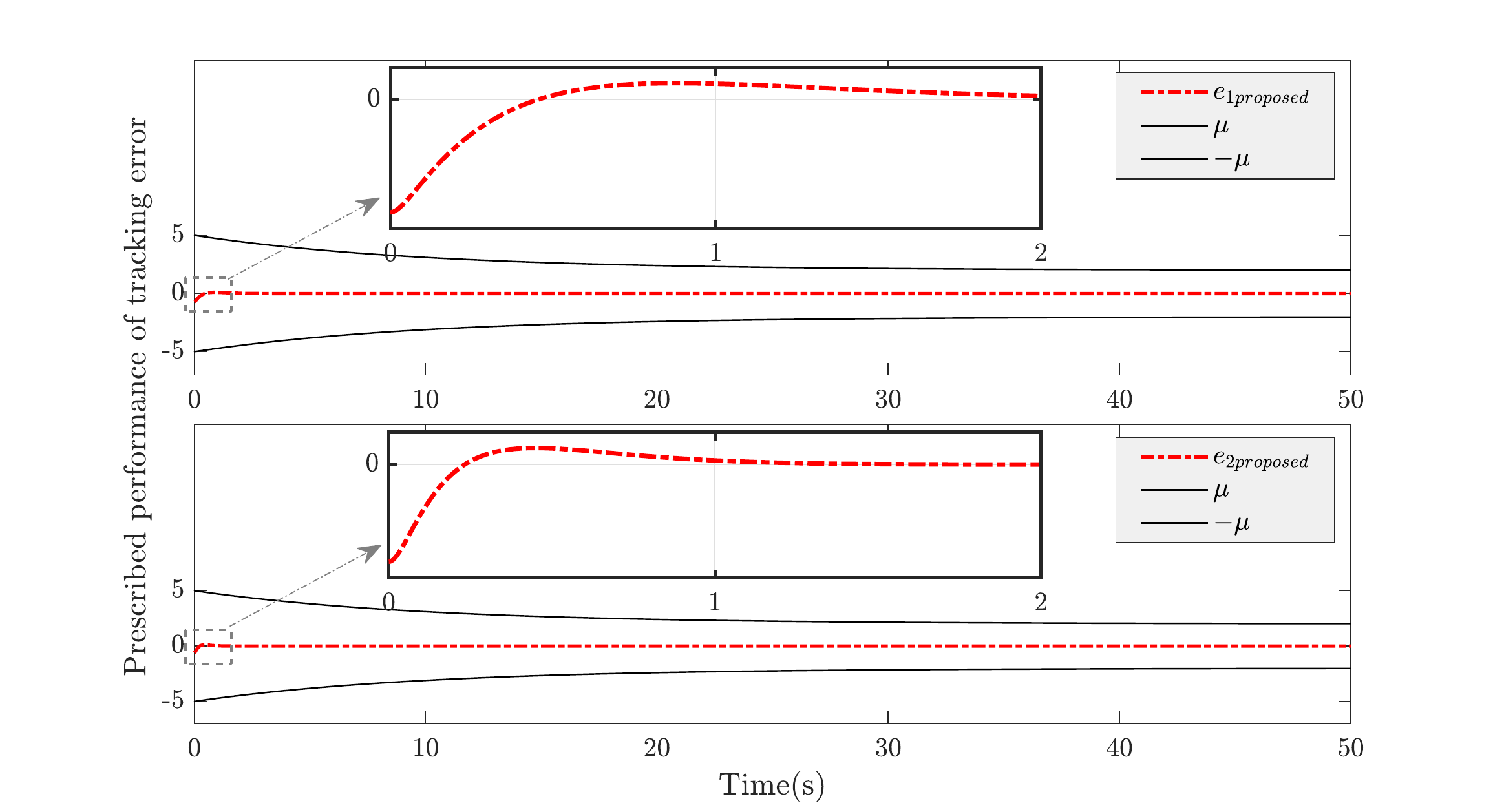}
    \caption{Tracking error within the bounds of the performance prescription function with $\mu_0=5$, $\mu_\infty$ = 2, and $l=0.1$. }
    \label{fig:ppf1}
\end{figure}

\begin{figure}[t]
    \centering
    \includegraphics[width = \linewidth]{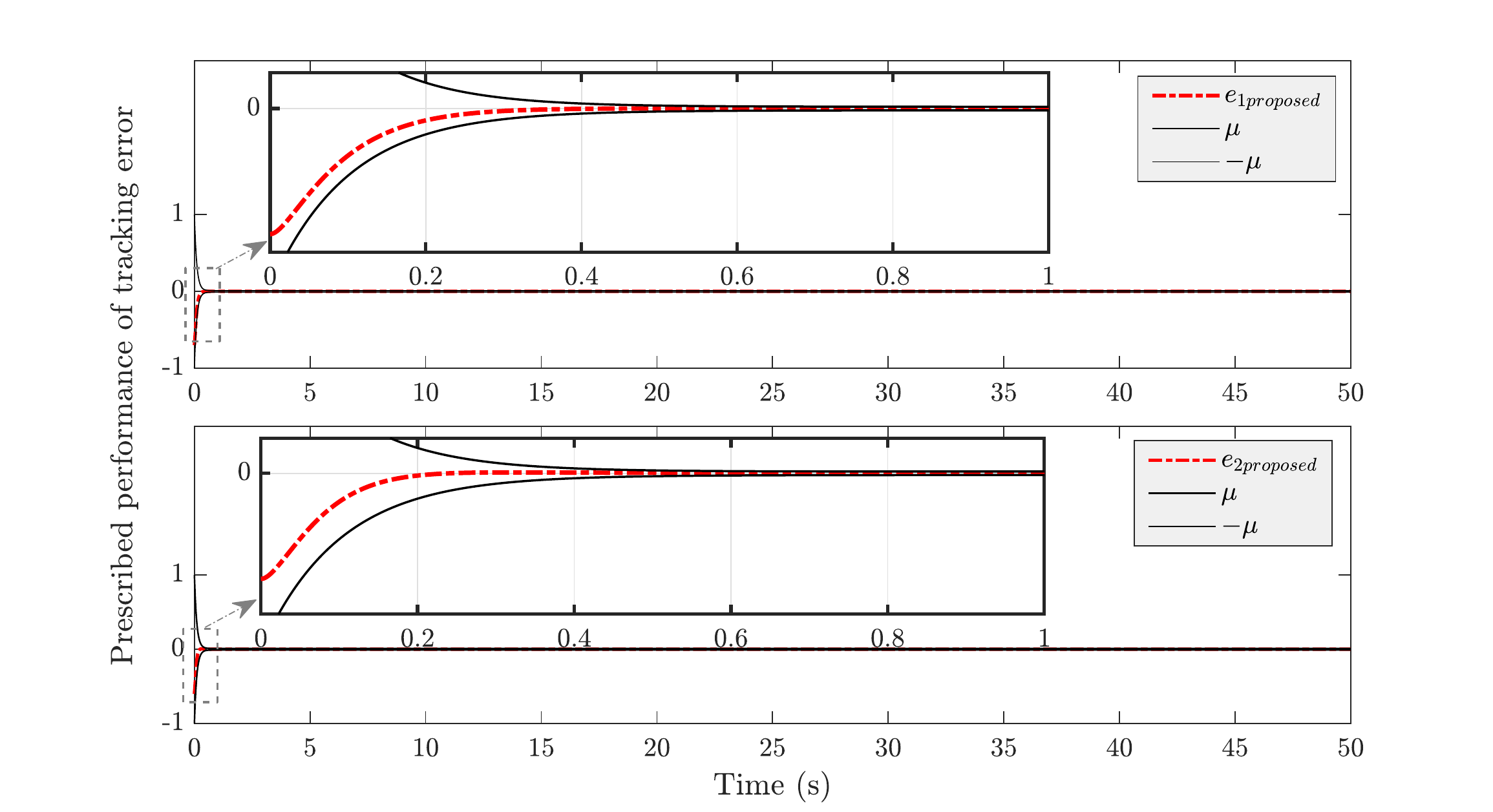}
    \caption{Tracking error within the bounds of the performance prescription function $\mu_0=1$, $\mu_\infty$ = 0.01, and $l=10$.}
    \label{fig:ppf2}
\end{figure}

Before the comparison, we would like to highlight the benefits of performance prescription. 
Figures \ref{fig:ppf1} and \ref{fig:ppf2} show the tracking error of the robot controlled by our method in nominal conditions. 
Two different performance prescription functions are considered, one with $\mu_0=5$, $\mu_\infty$ = 2, and $l=0.1$ (Fig. \ref{fig:ppf1}), and the other with $\mu_0=1$, $\mu_\infty$ = 0.01 and $l=10$ (Fig. \ref{fig:ppf2}).
The rest of the design parameters are set according to \ref{appendix2}, and the robot joints are tasked to follow sinusoidal trajectories.
It is evident that in both cases the tracking error is bounded within the funnels set by the prescription functions.
Clearly, setting tighter bounds in Fig. \ref{fig:ppf2} has not violated the boundaries and the controller has managed to keep $\|e_i\| \le 0.01$ with a fast decay rate. 
This shows the degree of flexibility that our method offers in prescribing the tracking error performance, a feature that is missing from the benchmark methods that will be discussed shortly and also from many other methods in fault-tolerant control literature.

Let us now compare our method with methods proposed in \cite{gao2021elm} and \cite{ma2016simultaneous}.
We set the design parameters for our controller as mentioned above. 
For the method by Gao et al., we used the same parameters that were proposed in \cite{gao2021elm}.
For the method by Ma et al., there was a need to fine-tune the parameters.
After several simulation trials, we chose the design parameters according to the original paper \cite{ma2016simultaneous} with the exceptions of $G_1 = 8$ and $\theta = 12$.
Furthermore, we set the initial conditions of all adaptation laws to zero.
We tasked the robot joints to follow sinusoidal trajectories and considered the following fault on the first angular position sensor with $E=\left(1\;0\right)^T$, and 
\begin{equation}
f = \left\{ {\begin{array}{*{20}{l}}
{0,}&{t < 25}\\
{3\tanh \left( {\frac{{\left(t - 25\right)^2}}{{0.2}}} \right),}&{t \ge 25}
\end{array}} \right. .
\end{equation}

\begin{figure}[t]
    \centering
    \includegraphics[width=\linewidth]{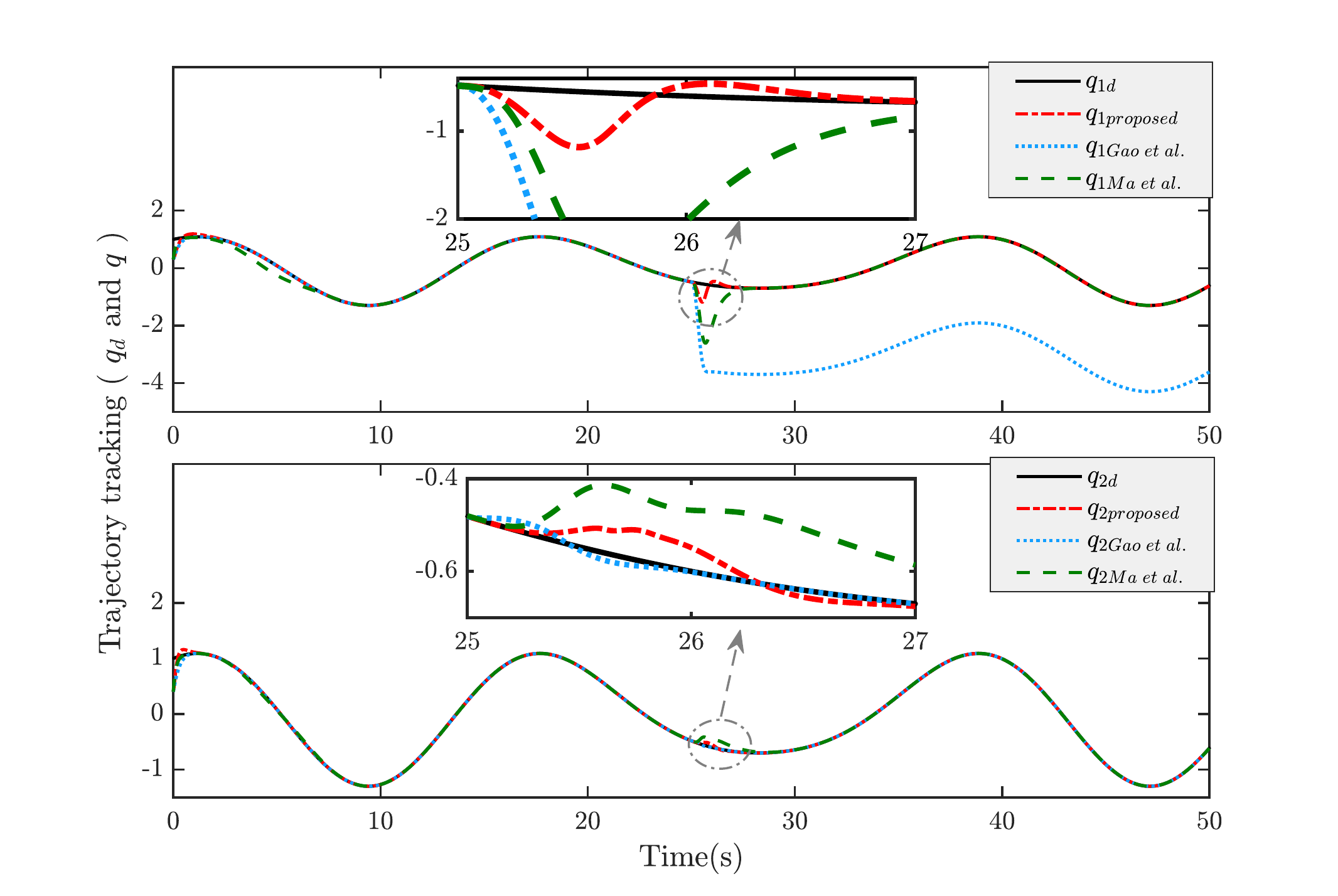}
    \caption{Time trajectory of robot joint positions for the proposed method, Gao et al. \cite{gao2021elm} and Ma et al. \cite{ma2016simultaneous}}
    \label{fig:positions}
\end{figure}

\begin{figure}[t]
    \centering
    \includegraphics[width=\linewidth]{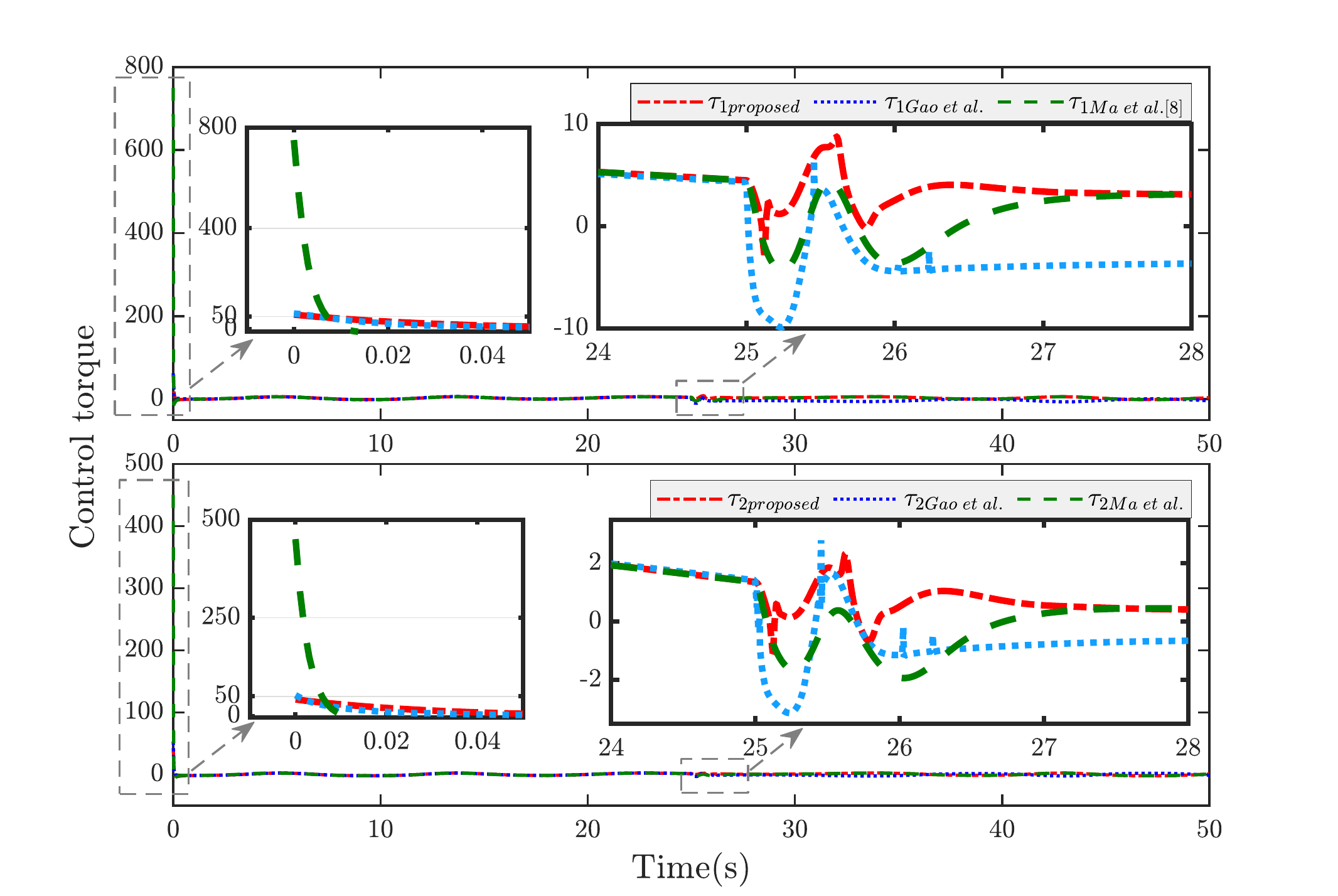}
    \caption{Control input for the proposed method, Gao et al. \cite{gao2021elm} and Ma et al. \cite{ma2016simultaneous}}
    \label{fig:torques}
\end{figure}

\begin{figure}[t]
    \centering
    \includegraphics[scale=0.75]{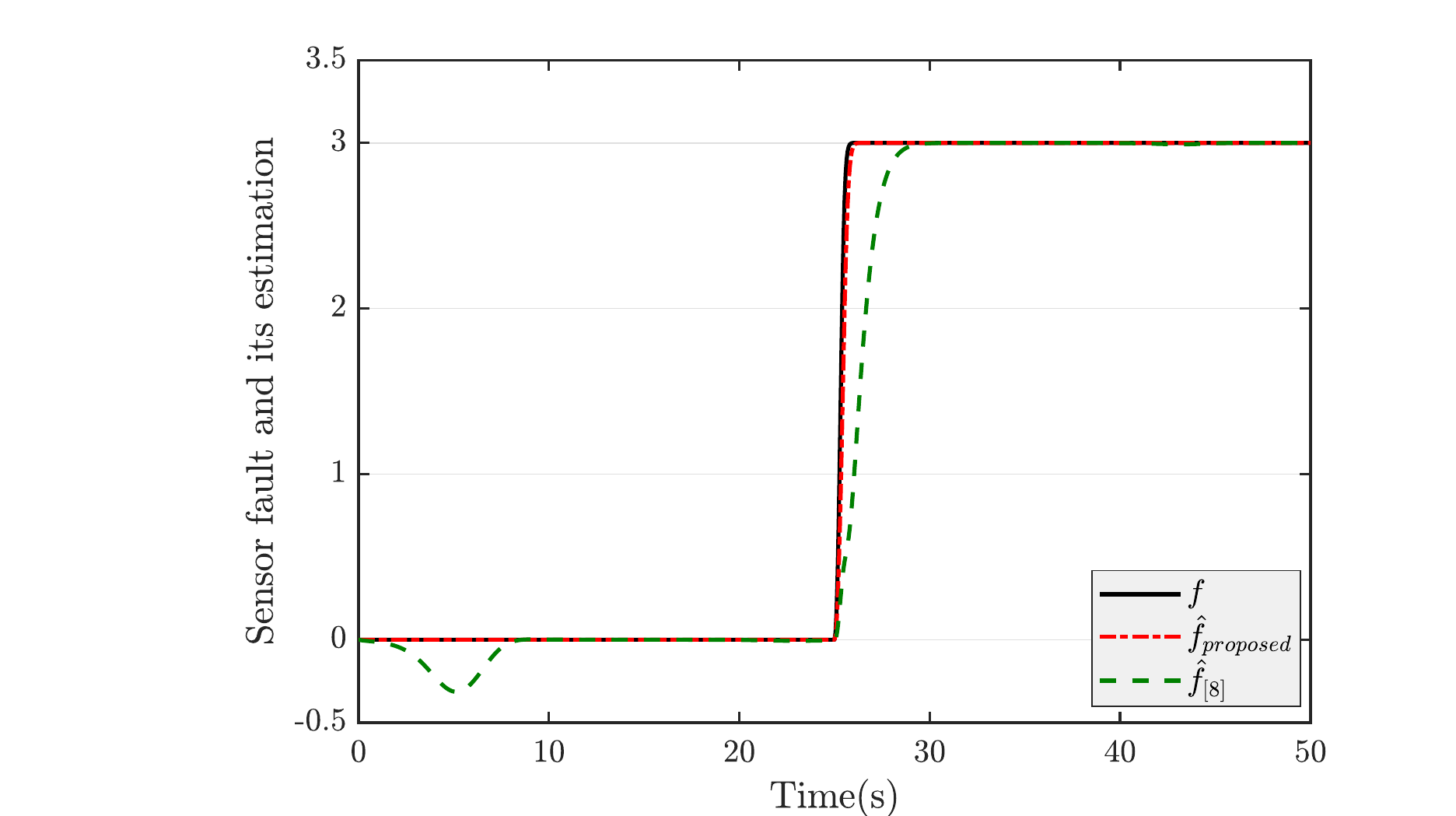}
    \caption{Fault estimation for the proposed method and Ma et al. \cite{ma2016simultaneous}}
    \label{fig:fault}
\end{figure}

Figures \ref{fig:positions} and \ref{fig:torques} demonstrate the trajectory-tracking performance and control inputs of all three methods.
When the sensor fault occurs at $t = 25\;s$, Gao et al.'s method fails to compensate for the fault's effect and undergoes a large steady-state error.
This observation is aligned with our discussion in Section \ref{se:tsmdsubsection}.
Our second-order sliding mode surface which was designed to address this steady-state error can effectively compensate for the effect of sensor fault and maintain accurate tracking.
Note that  Ma et al.'s method is also capable of compensating for the fault effect, but with a larger overshoot and slower convergence rate.
The superiority of our method partially stems from our adaptive observer design which is able to estimate the fault effects significantly faster than the Ma et al. method as shown in Fig. \ref{fig:fault}. 

Overall, the above simulation results show that our method (i) provides the ability to prescribe the performance of the trajectory-tracking error, (ii) addresses the steady-state error issue present in the Gao et al. method while maintaining similar fast convergence, and (iii) enables faster detection of fault and subsequently more effective fault compensation compared to the Ma et al. method.

\subsection{Example 2: Three-Degrees-of-Freedom Robotic Manipulator}
To further examine the effectiveness of the proposed method, we consider another example in which the trajectory control of a three-degrees-of-freedom manipulator is considered. 
This manipulator is a newly designed system to be used as a solar tracker base, as shown in Fig. \ref{fig:solarTracker}.
The model of the manipulator is given in \ref{appendix3}. More details can be found in \cite{ebrahimi2023identification}.

\begin{figure}[t]
    \centering
    \includegraphics[scale=0.5]{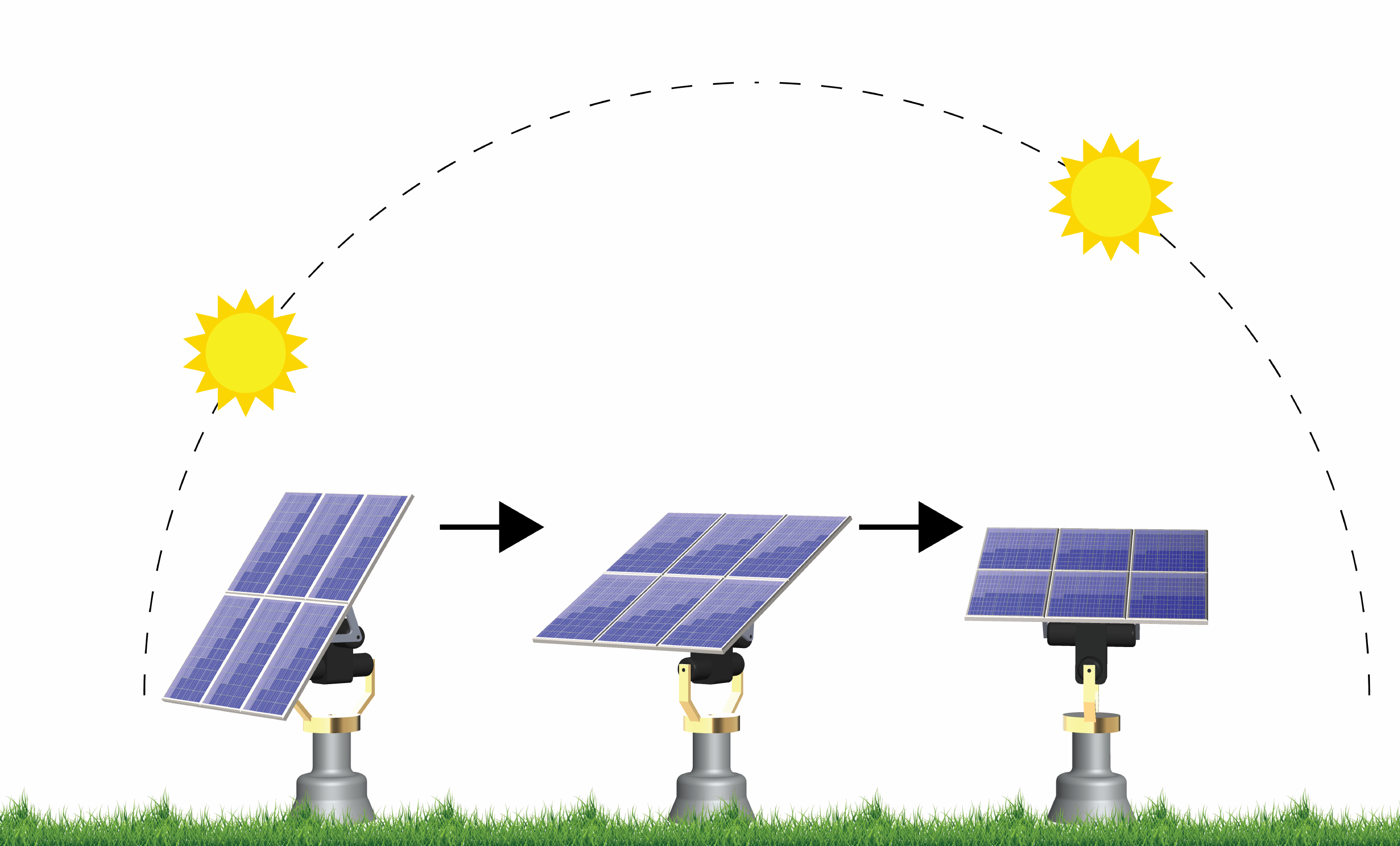}
    \caption{Solar tracker system}
    \label{fig:solarTracker}
\end{figure}

We study the system performance under different forms of sensor faults: (i) sinusoidal in the form of ${0.5\left( {\sin t + \sin 3.5t} \right)}\;rad$, (ii) an offset of $0.5\;rad$ with additive noise, (iii) a continuously increasing fault represented by a ramp signal, and (iv) a complicated fault scenario comprised of step, ramp, and sinusoidal signals.
We apply all these faults to the position sensor of all three joints at $t=25\;s$.
As mentioned in Section \ref{se:Intro}, one feature of sliding mode controllers is their inherent capability to passively deal with limited actuator faults. Although actuator fault is not the focus of this paper, to briefly explore the actuator fault tolerance of the proposed TSMC law, we introduce the following additive faults $5\sin\left(10\left(t-35\right)\right)$, $5\sin\left(5\left(t-35\right)\right)$, and $5\sin\left(7\left(t-35\right)\right)$ to the actuators of the three joints in all subsequent simulations.
We set the design parameters according to the values given in \ref{appendix3}.
Figure \ref{fig:solarTrackingError} presents the tracking error of all three joints under different faults. 
It is evident that the controller has compensated for actuator faults. With the occurrence of sensor faults, the tracking error degrades momentarily; however, our proposed fault estimation and control scheme manages to recover the system states and restore zero steady-state error.
Of note, even in the transient phase, the estimation error remains within the prescribed funnels.

\begin{figure}[t]
    \centering
    \includegraphics[width=\linewidth]{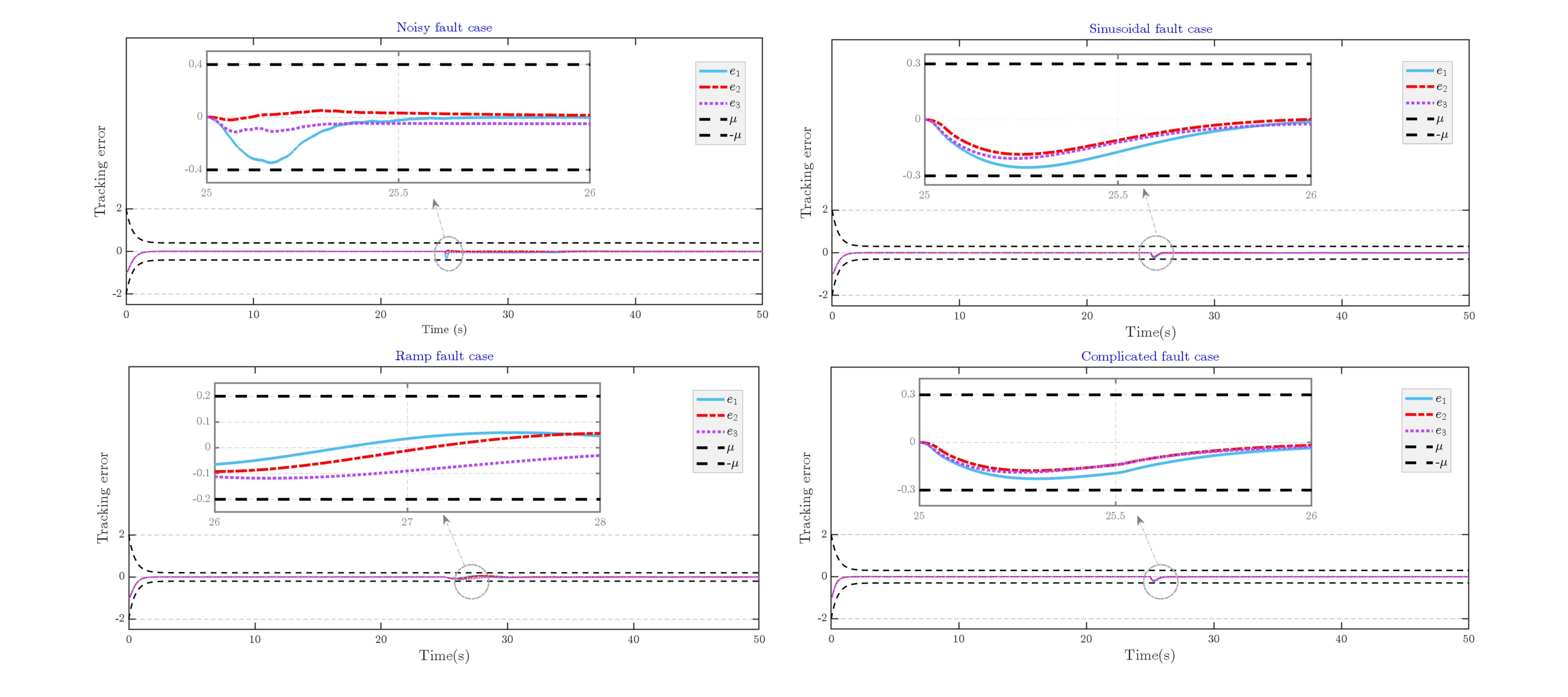}
    \caption{Tracking error in different fault scenarios}
    \label{fig:solarTrackingError}
\end{figure}

\begin{figure}[t]
    \centering
    \includegraphics[width=\linewidth]{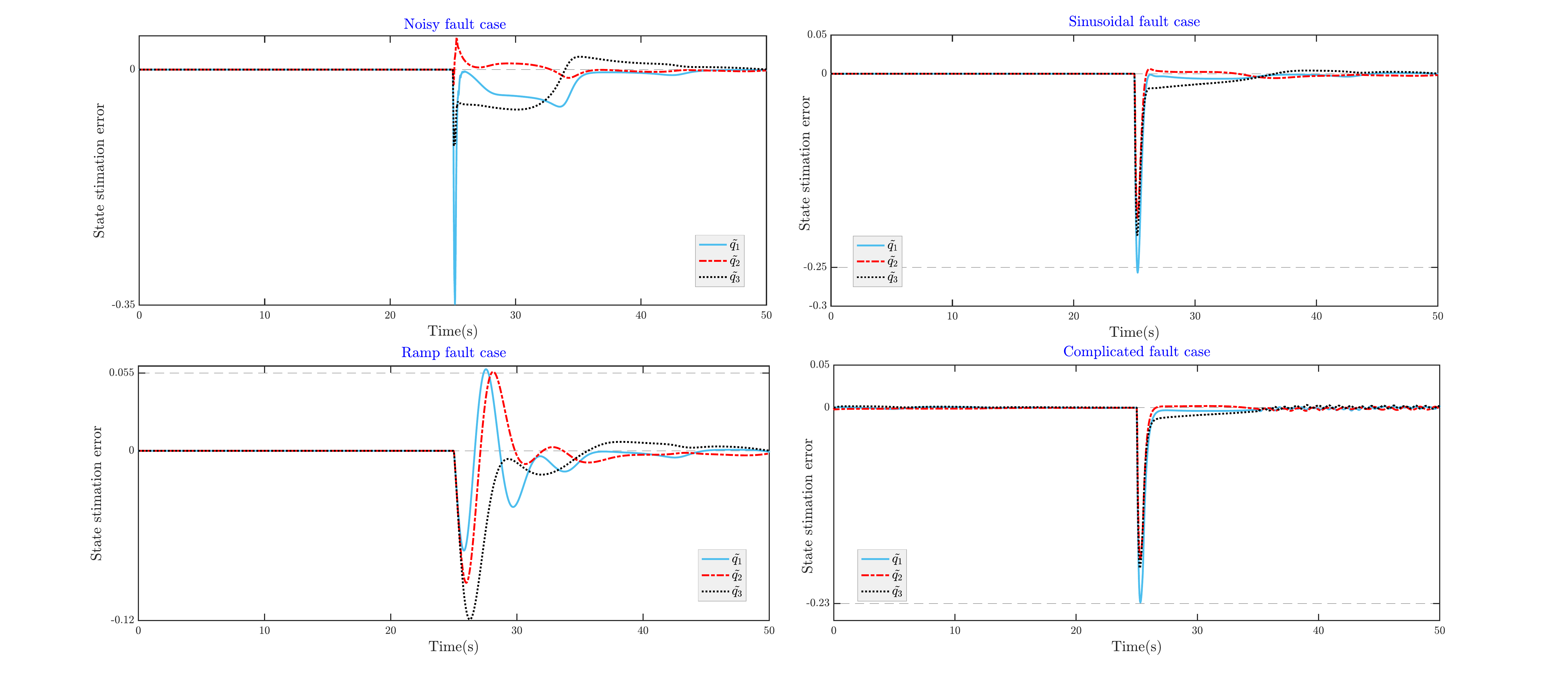}
    \caption{State estimation error in different fault scenarios}
    \label{fig:solarEstimationError}
\end{figure}

\begin{figure}[t]
    \centering
    \includegraphics[width=\linewidth]{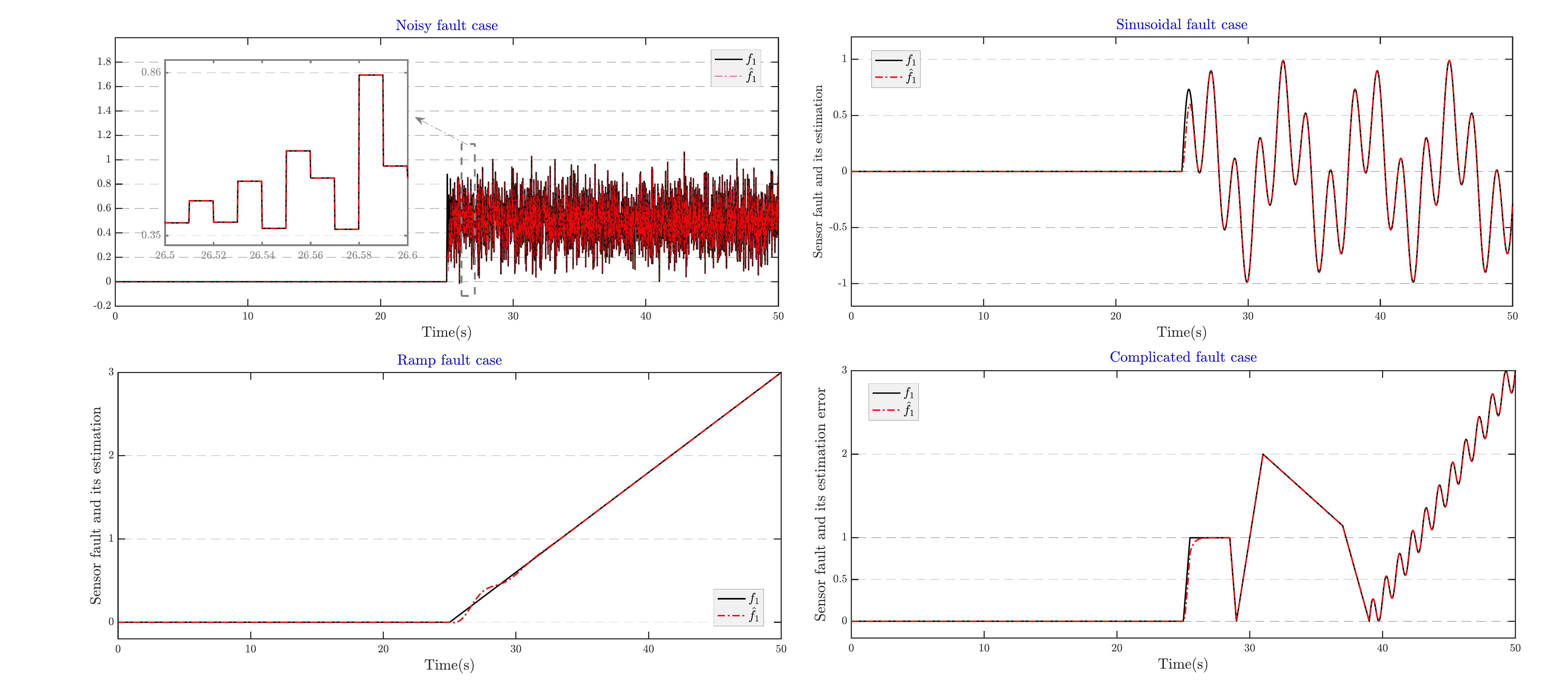}
    \caption{Sensor fault and its estimation error in different fault scenarios}
    \label{fig:solarFaultEstimation}
\end{figure}

\begin{figure}[t]
    \centering
    \includegraphics[width=\textwidth]{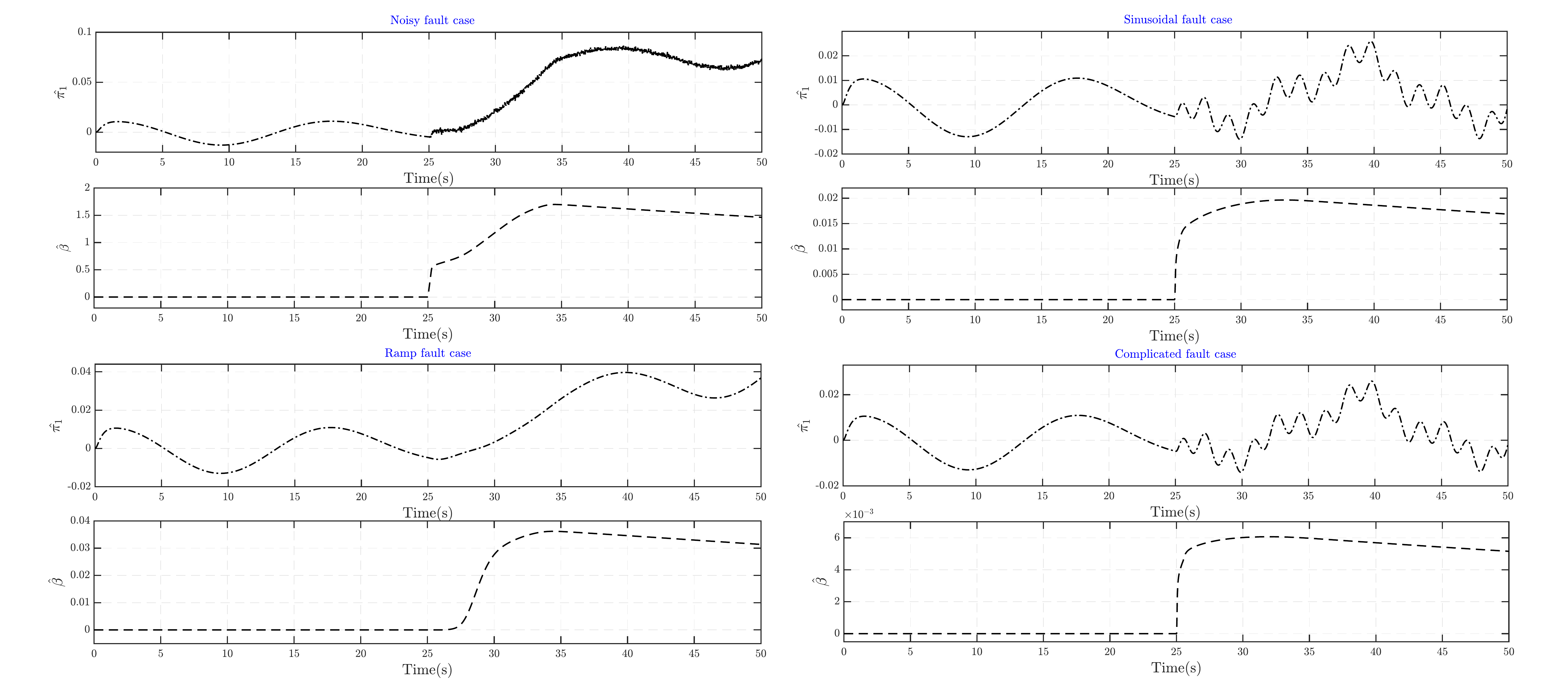}
    \caption{Time response of $\hat{\pi}_1$ and $\hat {\beta}_1$ in different fault scenarios}
    \label{fig:solarAdaptiveParameters}
\end{figure}

Figure \ref{fig:solarEstimationError} illustrates the state estimation error in each simulation scenario, showing the proposed observer's effectiveness to recover system states shortly after the occurrence of faults. 
This performance is attributed to the ability of the observer to converge to the actual fault values shown in Fig. \ref{fig:solarFaultEstimation}. 
The fault estimation part of the observer relies on two adaptive parameters $\hat \pi$ and $\hat \beta$ whose time trajectories are shown in Fig. \ref{fig:solarAdaptiveParameters}.
Note that since the results for different joints in each scenario are similar, we only present the results for the first joint.

\section{Conclusion}
This paper presented new results for sensor fault detection and compensation in robotic manipulators.
Our Lyapunov-based stability analysis and simulation experiments verified the proposed method both theoretically and numerically.
Our results conclude that the representation of sensor faults as virtual actuator faults combined with adaptive observer design given in \eqref{eq:observer} can be a viable technique for sensor fault detection.
In addition, the new TSMC law given in \eqref{eq:u} proved to be effective in compensating for the sensor fault effects.

The strengths of the proposed method include the ability to detect faults without the need to impose known bounds on the fault value or its derivative, and also fault compensation with a fast and fixed-time transient response, and the ability to prescribe system performance.
The above is achieved with only joint position measurements, despite many existing methods that require a measure of joint velocities in addition to position measurements.

Future research directions on the proposed method include the extension of the results to under-actuated systems, the incorporation of delays, input saturation, and/or dynamic uncertainty.
For the TSMC law development, our focus was to improve upon the tracking error of the method proposed in \cite{gao2021elm} while maintaining a similar convergence speed. However, the use of non-Lipschitz functions in the control law may lead to singularity issues and oscillatory behavior of the controller.
Addressing this issue can be a valuable extension of the current work. 
Additionally, the actuator fault tolerance properties of the TSMC law can be further explored and combined with active actuator FTC methods.
Another direction to extend this work is to consider fault-tolerant 
cooperative control problems where multiple robotic manipulators collaborate to accomplish a task. In this context, there exist interesting observer-based results such as \cite{doostmohammadian2023sensor, doostmohammadian2022distributed} that can be leveraged to extend the current work.

\appendix
\section{Mathematical Background}
\subsection{Useful Inequalities}

\textbf{Lemma 1 - Young's Inequality}\label{lemma1} \cite{wang2020cooperative}: \textit{For any given $a, b \in \mathbb{R}^n$ we have}
\begin{equation}\label{eq:young}
    2a^TSQb \le a^TSPS^Ta+b^TQ^TP^{-1}Qb,
\end{equation}
\textit{where $P>0$, $S$, and $Q$ have appropriate dimensions.}

\textbf{Lemma 2}\label{lemma2} \cite{zuo2015non}: \textit{For $v = \left( {{v_1},{v_2}, \cdots ,{v_n}} \right)^T \in \mathbb{R}^n$, $v_i > 0$, and the constants $0<a_1<1$ and $a_2>1$, we have}
\begin{equation}\label{eq:lemma2-1}
    \sum_{i=1}^{n}v_{i}^{a_1}\geq\left(\sum_{i=1}^{n}v_i\right)^{a_1},
\end{equation}
\textit{and}
\begin{equation}\label{eq:lemma2-2}
    \sum_{i=1}^{n}v_{i}^{a_2}\geq n^{1-a_2}\left(\sum_{i=1}^{n}v_i\right)^{a_2}.
\end{equation}

\textbf{Lemma 3}\label{lemma3} \cite{dastres2020neuralB}: \textit{Given two positive scalars $a$ and $b$, we have}
\begin{equation}
    \left|a\right|\ge 0.8814b \rightarrow 1-2\tanh^2\left(\frac{a}{b}\right) \le 0,
\end{equation}
\textit{and}
\begin{equation}
    \left|a\right| < 0.8814b \rightarrow 0<1-2\tanh^2\left(\frac{a}{b}\right) < 1.
\end{equation}

\subsection{Ultimate Boundedness}
\textbf{Lemma 4:}\label{lemma4} \cite{krstic1995nonlinear} \textit{Let $V$ and $\rho$ be real-valued positive definite functions, and let $\alpha$ and $\beta$ be positive constants.
If they satisfy the differential inequality
\begin{equation}
    \dot V \le -\alpha V + \beta \rho^2,\;\;\;v(0) \ge 0,
\end{equation}
then we have
\begin{equation}
    V(t) \le V(0)e^{-\alpha t} + \beta \int_0^t e^{-\alpha(t-\tau)}\rho(\tau)^2 d \tau.
\end{equation}}

\subsection{Fixed-Time Stability}
\textbf{Lemma 5}\label{lemma5} \cite{jiang2016fixed}: \textit{Consider the following nonlinear system
\begin{equation}\label{eq:def1}
    \dot{\chi}\left(t\right)=f\left(\chi\left(t\right)\right),\;\;\;\chi\left(0\right)=\chi_0,
\end{equation}
where $\chi\in\mathbb{R}^n$, and $f\left(\chi\left(t\right)\right):\mathbb{R}^n\ \rightarrow\ \mathbb{R}^n$ is a continuous function.
The system \eqref{eq:def1} is said to be fixed-time stable if there exists a continuous positive definite function $V\left(\chi\right)$ such that 
\begin{equation}
    \dot{V}\left(\chi\right)\le-a\ V\left(\chi\right)^\alpha-b\ V\left(\chi\right)^\vartheta+\zeta,
\end{equation}
where $a>0$, $b>0$, $0<\alpha<1$, $\vartheta > 1$, and $0<\zeta<\infty$.
The convergence region is 
\begin{equation}
    \Delta=\left\{\chi|\ V\left(\chi\right)\le\min{\left\{\left(\frac{\zeta}{\left(1-\theta\right)a}\right)^\frac{1}{\alpha},\left(\frac{\zeta}{\left(1-\theta\right)b}\right)^\frac{1}{\vartheta}\right\}}\right\},
\end{equation}
where $0<\theta<1$, and the settling time is $T\left(\chi_0\right)$ such that $T\left(\chi_0\right)<T_{max}$, and $0<T_{max}\le\frac{1}{a\left(1-\alpha\right)}+\frac{1}{b\left(1-\vartheta\right)}$.}

\textbf{Lemma 6}\label{lemma6} \cite{gao2021elm}:\textit{Consider the following scalar system}
\begin{equation}\label{eq:lemma6-1}
    \dot{\chi}=-\frac{1}{\varphi\left(\chi\right)}\left(\underline{\lambda}{\rm{sgn}}^{p^\ast}\left(\chi\right)+\bar{\lambda}{\rm{sgn}}^\frac{\bar{p}}{\bar{q}}\left(\chi\right)\right),
\end{equation}
\textit{where $\varphi \left( \chi  \right) = {a_1} + \left( {1 - {a_1}} \right){e^{ - {b_1}{{\left| \chi  \right|}^{{c_1}}}}}$, ${p^*} = 0.5\left( {{{\underline p} \mathord{\left/
 {\vphantom {{\underline p} {\underline q}}} \right.
 \kern-\nulldelimiterspace} {\underline q}} + \left( {{{\underline p} \mathord{\left/
 {\vphantom {{\underline p} {\underline q}}} \right.
 \kern-\nulldelimiterspace} {\underline q}} - 1} \right){\rm{sgn}}\left( {\left| \chi  \right| - 1} \right)} \right)$, $\underline \lambda >0$, $\bar \lambda > 0$, $0 < a_1 < 1$, $ b_1 >0$, $c_1$ is a positive even integer, and $\underline p > 0$, $\underline q > 0$, $\bar p > 0$, $\bar q > 0$, $\underline p > \underline q$, and $\bar p < \bar q$ are odd integers. 
The system \eqref{eq:lemma6-1} is fixed-time stable with the following convergence time
\begin{equation}\label{eq:convergenceTime}
    T_{s1}\left(\chi_0\right)<\frac{\underline{q}}{\underline{\lambda}\left(\underline{p}-\underline{q}\right)}+\frac{\bar{q}}{\bar{q}-\bar{p}}\frac{1}{\underline{\lambda}}\ln{\left(1+\frac{\underline{\lambda}}{\bar{\lambda}}\right)}.
\end{equation}
}

\section{Parameters Values Used in Example 1 of Simulations}\label{appendix2}
The robotic manipulator's parameters are set according to \cite{gao2021elm}.

The controller parameters are set as $\underline{\lambda}=1$, $\bar{\lambda}=2$,
$a=0.7$, $b=1$, $c=2$, $\underline{p}=\bar{q}=p_{\sigma}=\bar{q}_{\sigma}=25$, $\underline{q}=\bar{p}=\bar{p}_\sigma=q_\sigma=23$, and  $c_{1\sigma}=\bar{c}_{1\sigma}=k_{1\sigma}=10$.

The performance prescription function parameters are set as $\mu_0=5$, $\mu_\infty$ = 2, and $l=0.1$ for Fig. \ref{fig:ppf1}, and $\mu_0=1$, $\mu_\infty$ = 0.01, and $l=10$ for Fig. \ref{fig:ppf2}.

The observer parameters are set as $\rho=100$, $\Gamma=0.05$, $\gamma=1$, $\varepsilon=I_2$, $\rho_c=0.001$,
\begin{equation*}
    A_3 = \left(\begin{matrix}20 & 0.1\\0.1 & 20\end{matrix}\right),
\end{equation*}
\begin{equation*}
    P = \left(\begin{matrix}
5.9419&0.0001&-0.6670&0.0001&-1.9084&0.0040\\
0.0001&5.9419&0.0001&-0.6670&0.0040&-1.9084\\
-0.6670&0.0001&0.6044&-0.0000&-0.0265&0.0001\\
0.0001&-0.6670&-0.0000&0.6044&0.0001&-0.0265\\
-1.9084&0.0040&-0.0265&0.0001&4.7186&-0.0106\\
0.0040&-1.9084&0.0001&-0.0265&-0.0106&4.7186   
    \end{matrix}\right),
\end{equation*}
and
\begin{equation*}L=
    \left(\begin{matrix}
        14.1213&0.0036\\
        0.0536&14.1214\\
        15.2936&0.0015\\
        0.0626&15.2936\\
        -7.2526&-0.1340\\
        0.0006&-7.2528
    \end{matrix}\right).
\end{equation*}

\section{Parameters Values Used in Example 2 of Simulations}\label{appendix3}
The elements of the $M(q)$ matrix include 
\begin{equation*}
\begin{array}{l}
{m_{11}} = s_2^2\left( {{m_2}\ell _2^2 + {m_3}{{\left( {{c_3}{\ell _3} + {L_2}} \right)}^2} + {I_{{y_2}}} + {I_{{y_3}}}} \right) + {m_3}s_3^2\ell _3^2 + {I_{{z_1}}}\\
\;\;\;\;\;\;\;\; + c_2^2\left( {{I_{{z_2}}} + s_3^2{I_{{x_3}}} + c_3^2{I_{{z_3}}}} \right),
\end{array}
\end{equation*}
\begin{equation*}
{m_{12}} = {m_{21}} = {s_3}{c_2}\left( {{c_3}\left( {{I_{{x_3}}} - {I_{{z_3}}}} \right) - {m_3}{\ell _3}\left( {{c_3}{\ell _3} + {L_2}} \right)} \right),
\end{equation*}
\begin{equation*}
    {m_{13}} = {m_{31}} = {s_2}\left( {{m_3}{\ell _3}\left( {{\ell _3} + {c_3}{L_2}} \right) - {I_{{y_3}}}} \right),
\end{equation*}
\begin{equation*}
    m_{22}=m_2\ell_2^2+m_3\left(c_3\ell_3+L_2\right)^2+I_{x_2}+c_3^2I_{x_3}+s_3^2I_{z_3}
\end{equation*}
\begin{equation*}
    m_{23}=m_{32}=0,
\end{equation*}
and
\begin{equation*}
    m_{33}=m_3l_3^2+I_{y_3}.
\end{equation*}
The elements of the $D(q,\dot q)$ include
\begin{equation*}
    \begin{array}{l}
{d_{11}} = \left( {{s_3}{c_3}\left( {{m_3}l_3^2 + c_2^2({I_{{x_3}}} - {I_{{z_3}}}} \right) - s_2^2\left( {{m_3}{s_3}{\ell _3}\left( {{c_3}{\ell _3} + {L_2}} \right)} \right)} \right){{\dot q}_3}\\
\;\;\;\;\;\; + {s_2}{c_2}\left( {{m_2}\ell _2^2 + {m_3}{{\left( {{c_3}{\ell _3} + {L_2}} \right)}^2} + {I_{{y_2}}} + {I_{{y_3}}} - {I_{{z_2}}} - s_3^2{I_{{x_3}}} - c_3^2{I_{{z_3}}}} \right){{\dot q}_2},
\end{array}
\end{equation*}
\begin{equation*}
    \begin{array}{l}
{d_{12}} = {s_2}{c_2}\left( {{m_2}\ell _2^2 + {m_3}{{\left( {{c_3}{\ell _3} + {L_2}} \right)}^2} + {I_{{y_2}}} + {I_{{y_3}}} - {I_{{z_2}}} - s_3^2{I_{{x_3}}} - c_3^2{I_{{z_3}}}} \right){{\dot q}_1}\\
\;\;\;\;\;\;\; - {s_2}{s_3}\left( {{c_3}\left( {{I_{{x_3}}} - {I_{{z_3}}}} \right) - {m_3}{\ell _3}\left( {{c_3}{\ell _3} + {L_2}} \right)} \right){{\dot q}_2}\\
\;\;\;\;\;\;\; + {c_2}\left( {{m_3}s_3^2\ell _3^2 + c_3^2\left( {{I_{{x_3}}} - {I_{{z_3}}}} \right) + \frac{1}{2}\left( {{I_{{z_3}}} - {I_{{y_3}}} - {I_{{x_3}}}} \right)} \right){{\dot q}_3},
\end{array}
\end{equation*}
\begin{equation*}
    \begin{array}{l}
{d_{13}} = \left( {{s_3}{c_3}({m_3}l_3^2 + c_2^2({I_{{x_3}}} - {I_{{z_3}}})) - s_2^2\left( {{m_3}{s_3}{\ell _3}\left( {{c_3}{\ell _3} + {L_2}} \right)} \right)} \right){{\dot q}_1}\\
\;\;\;\;\;\;\; + {c_2}\left( {{m_3}s_3^2\ell _3^2 + c_3^2\left( {{I_{{x_3}}} - {I_{{z_3}}}} \right) + \frac{1}{2}\left( {{I_{{z_3}}} - {I_{{x_3}}} - {I_{{y_3}}}} \right)} \right){{\dot q}_2}\\
\;\;\;\;\;\;\; - {m_3}{s_2}{s_3}{\ell _3}{L_2}{{\dot q}_3},
\end{array}
\end{equation*}
\begin{equation*}
    \begin{array}{l}
{d_{21}} =  - {s_2}{c_2}\left( {{m_2}\ell _2^2 + {m_3}{{\left( {{c_3}{\ell _3} + {L_2}} \right)}^2} + {I_{{y_2}}} + {I_{{y_3}}} - {I_{{z_2}}} - s_3^2{I_{{x_3}}} - c_3^2{I_{{z_3}}}} \right){{\dot q}_1}\\
\;\;\;\;\;\;\; + {c_2}\left( {c_3^2\left( {{I_{{x_3}}} - {I_{{z_3}}}} \right) - {m_3}{c_3}{\ell _3}\left( {{c_3}{\ell _3} + {L_2}} \right) + \frac{1}{2}\left( {{I_{{y_3}}} + {I_{{z_3}}} - {I_{{x_3}}}} \right)} \right){{\dot q}_3},
\end{array}
\end{equation*}
\begin{equation*}
    c_{22}=s_3\left(c_3\left(I_{z_3}-I_{x_3}\right)-m_3l_3\left(c_3l_3+L_2\right)\right)\dot{q_3}
\end{equation*}
\begin{equation*}
    \begin{array}{l}
{d_{23}} = {c_2}\left( {c_3^2\left( {{I_{{x_3}}} - {I_{{z_3}}}} \right) - {m_3}{c_3}{\ell _3}\left( {{c_3}{\ell _3} + {L_2}} \right) + \frac{1}{2}\left( {{I_{{y_3}}} + {I_{{z_3}}} - {I_{{x_3}}}} \right)} \right){{\dot q}_1}\\
\;\;\;\;\;\; + {s_3}\left( {{c_3}\left( {{I_{{z_3}}} - {I_{{x_3}}}} \right) - {m_3}{l_3}\left( {{c_3}{l_3} + {L_2}} \right)\;} \right){{\dot q}_2},
\end{array}
\end{equation*}
\begin{equation*}
    \begin{array}{l}
{d_{31}} = \left( {s_2^2\left( {{m_3}{s_3}{\ell _3}\left( {{c_3}{\ell _3} + {L_2}} \right)} \right) - {s_3}{c_3}\left( {{m_3}l_3^2 + c_2^2\;\left( {{I_{{x_3}}} - {I_{{z_3}}}} \right)} \right)} \right){{\dot q}_1}\\
\;\;\;\;\;\; + {c_2}\;\left( {{m_3}{c_3}{\ell _3}\left( {{c_3}{\ell _3} + {L_2}} \right) - c_3^2\left( {{I_{{x_3}}} - {I_{{z_3}}}} \right) - \frac{1}{2}\left( {{I_{{y_3}}} + {I_{{z_3}}} - {I_{{x_3}}}} \right)} \right){{\dot q}_2},
\end{array}
\end{equation*}
\begin{equation*}
    \begin{array}{l}
{d_{32}} = {c_2}\left( {{m_3}{c_3}{\ell _3}\left( {{c_3}{\ell _3} + {L_2}} \right) - c_3^2\left( {{I_{{x_3}}} - {I_{{z_3}}}} \right) - \frac{1}{2}\left( {{I_{{y_3}}} + {I_{{z_3}}} - {I_{{x_3}}}} \right)} \right){{\dot q}_1}\\
\;\;\;\;\;\;\; + {s_3}\left( {{m_3}{l_3}\left( {{c_3}{l_3} + {L_2}} \right) - {c_3}\left( {{I_{{z_3}}} - {I_{{x_3}}}} \right)} \right){{\dot q}_2},
\end{array}
\end{equation*}
and $d_{33}=0$, where $s_i$ and $c_i$ stand for $\sin(q_i)$ and $\cos(q_i)$, respectively.

The vector of the effect of gravitational force is expressed as
\begin{equation*}
    G\left( q \right) =  - g\left( {\begin{array}{*{20}{c}}
0\\
{{s_2}\left( {{m_2}{l_2} + {m_3}\left( {{c_3}{\ell _3} + {L_2}} \right)} \right)}\\
{{m_3}{l_3}{s_3}{c_2}}
\end{array}} \right).
\end{equation*}

The parameter values of the robot set as
$m_1 = 27.387\;kg$, $m_2 = 15.843\;kg$, $m_3 = 40.53\;kg$, $l_1 = 0.07\;m$, $l_2 = 0.085\;m$, $l_3 = 0.326\;m$, $L_1 = 0.410\;m$, $L_2 = 0.170\;m$, $L_3 = 0.5\;m$, $I_{x_1} = 0.285\;kg.m^{-2}$, $I_{x_2} = 0.254\;kg.m^{-2}$, $I_{x_3}= 2.161\;kg.m^{-2}$, $I_{y_1}=0.458\;kg.m^{-2}$, $I_{y_2}=0.254\;kg.m^{-2}$, $I_{y_3}=1.949\;kg.m^{-2}$, $I_{z_1}=0.427\;kg.m^{-2}$, $I_{z_2}=0.229\;kg.m^{-2}$, $I_{z_3}=3.341\;kg.m^{-2}$, and $g = 9.807 m.s^{-2}$.

The controller parameters are set as $\underline{\lambda} = 0.01$, $\bar{\lambda}=1$, $\underline{p}=\bar{q}=11$, $\underline{q}=\bar{p}=9$, $a=0.7$, $b=1$, $c=2$, $c_{1\sigma}=\bar{c}_{1\sigma}=k_{1\sigma}=1$, $p_\sigma=\bar{q}_\sigma=11$, and $\bar{p}_\sigma=q_\sigma = 9$.

The performance prescription function parameters are set as $l=2$, $\mu_0=2$, and $\mu_\infty=0.3$.

The observer parameters are set as $\rho=\rho_c=\Gamma=0.01$, $\gamma=10$, $\varepsilon=300I_3$, $A_3=100I_3$,
\begin{equation*}
\tiny P = 
    \left(\begin{matrix}
    3.23637&2.23e-19&-8.10e-17&-1.07894&-1.88e-16&4.07e-17&-0.01348&6.85e-17&-9.24e-17\\
    2.23e-19&3.23637&1.18e-15&3.68e-16&-1.07894&3.53e-16&-6.84e-17&-0.01348&-5.63e-16\\
    -8.10e-17&1.18e-15&3.23637&-4.51e-17&-3.99e-16&-1.07894&9.07e-17&5.86e-16&-0.01348\\
    -1.07894&3.68e-16&-4.51e-17&3.237968&-3.31e-17&-2.92e-17&-0.03236&-1.35e-17&7.44e-19\\
    -1.88e-16&-1.07894&-3.99e-16&-3.31e-17&3.237968&9.90e-16&3.71e-19&-0.03236&-1.45e-17\\
    4.07e-17&3.53e-16&-1.07894&-2.92e-17&9.90e-16&3.237968&8.63e-19&6.08e-18&-0.03236\\
    -0.01348&-6.84e-17&9.07e-17&-0.03236&3.71e-19&8.63e-19&2.696713&-1.54e-17&-2.94e-17\\
    6.85e-17&-0.01348&5.86e-16&-1.35e-17&-0.03236&6.08e-18&-1.54e-17&2.696713&2.98e-17\\
    -9.24e-17&-5.63e-16&-0.01348&7.44e-19&-1.45e-17&-0.03236&-2.94e-17&2.98e-17&2.696713\\
    \end{matrix}\right),
\end{equation*}
 and 
\begin{equation*}
    L = \left(\begin{matrix}
        93.74817&-2.37e-15&1.78e-15\\
        -2.23e-15&93.74817&-4.03e-14\\
        1.69e-15&-4.04e-14&93.74817\\
        31.24764&-1.11e-14&2.12e-15\\
        4.98e-15&31.24764&-1.19e-14\\
        -2.50e-16&-3.27e-14&31.24764\\
        -98.6566&2.23e-15&-3.08e-15\\
        -2.16e-15&-98.6566&-2.05e-14\\
        3.19e-15&1.95e-14&-98.6566
    \end{matrix}\right).
\end{equation*}

\bibliographystyle{elsarticle-num}
\bibliography{References}

\end{document}